\documentclass[num-refs]{wiley-article}

\usepackage{color}
\usepackage{ulem}

\usepackage{siunitx}
\usepackage{graphicx} 
\usepackage{booktabs}
\usepackage{float}
\usepackage{algorithm}
\usepackage{algpseudocode}
\usepackage{setspace}
\usepackage{mathrsfs}
\usepackage{multirow}
\usepackage{multicol}
\usepackage{subfig}
\papertype{Original Article}
\paperfield{Journal Section}

\title{Automatic Stack Velocity Picking Using an Unsupervised Ensemble Learning Method}

\author[1]{Hongtao~Wang}
\author[1]{Jiangshe~Zhang}
\author[1]{Chunxia~Zhang}
\author[1]{Li~Long}
\author[2]{Weifeng~Geng}

\affil[1]{School of Mathematics and Statistics, Xi’an Jiaotong University, Xi’an, Shaanxi, 710049, P.R.China}
\affil[2]{Geophysical Technology Research Center of Bureau of Geophysical Prospecting, Zhuozhou, Hebei, 072751, P.R.China}

\corraddress{Name: Jiangshe~Zhang\\Address: School of Mathematics and Statistics, Xi’an Jiaotong University, Xi’an, Shaanxi, 710049, P.R.China}
\corremail{jszhang@mail.xjtu.edu.cn}

\fundinginfo{The research is supported by the National Key Research and Development Program of China under grant 2020AAA0105601, the National Natural Science Foundation of China under grant 61976174 and 61877049.}

\runningauthor{Wang et al.}

\begin{document}

\maketitle

\begin{abstract}
Seismic velocity picking algorithms that are both accurate and efficient can greatly speed up seismic data processing, with the primary approach being the use of velocity spectra.
Despite the development of some supervised deep learning-based approaches to automatically pick the velocity, they often come with costly manual labeling expenses or lack interpretability.
In comparison, using physical knowledge to drive unsupervised learning techniques has the potential to solve this problem in an efficient manner.
We suggest an Unsupervised Ensemble Learning (UEL) approach to achieving a balance between reliance on labeled data and picking accuracy, with the aim of determining the stack velocity.
UEL makes use of the data from nearby velocity spectra and other known sources to help pick efficient and reasonable velocity points, which are acquired through a clustering technique.
Testing on both the synthetic and field data sets shows that UEL is more reliable and precise in auto-picking than traditional clustering-based techniques and the widely used Convolutional Neural Network (CNN) method.

\end{abstract}

\section{Introduction}
Stack velocity estimation from the prestack gather is critical in seismic data processing, which ensures that the following processes obtain an optimal poststack gather \citep{yilmaz2001seismic}. 
Stack velocity is used in Normal MoveOut (NMO) correction to flatten the prestack gather. Thus, stack velocity is also called NMO velocity. To manually estimate stack velocity more conveniently, \cite{taner1969velocity} proposed velocity spectrum, which is computed by some criterion \citep{neidell1971semblance}, e.g., semblance method. However, manual picking from the semblance velocity spectrum is not only a very time-consuming process but also requires an experienced processing analyst. Therefore, developing automatic picking methods is still critical to liberating productivity.

Most of the current stack velocity automatic picking methods are based on the semblance velocity spectrum, where the reasonable peaks with high coherence values are picked as stack velocities. Also, some researchers picked the stack velocity directly based on the prestack gather \citep{calderon1998automatic, zhang2016automatic, ma2018automatic, biswas2019estimating}. In addition, hybrid methods, combining both the information of velocity spectrum and other information, e.g. prestack gather and stacked gather slice were introduced to pick the stack velocity \citep{ferreira2020automatic, 2021Ma, wang2022automatic}. 
In the current paper, we will focus on discussing the automatic picking method based on the semblance velocity spectra. 
With the rapid development of machine learning methods in geophysics, 
various automatic picking methods based on machine learning or deep learning methods have been developed so far. Broadly speaking, they can be mainly divided into two kinds of methods: supervised methods and unsupervised methods. 

In the supervised methods, the models are driven by big data with manual labels, and most of them utilize successful deep learning models in Computer Vision (CV), e.g., Recurrent Neural Network (RNN) \citep{zaremba2014recurrent}, U-Net \citep{ronneberger2015u}, You Only Look Once (YOLO) \citep{7780460}, etc. The insights of these automatic picking models are natural, that is, mapping the velocity spectrum to a few points or values. Subsequently, the task can be reduced to an image classification problem, an object detection problem, or a semantic segmentation task. First, CNN models were used to classify both the local slices and the global image of the velocity spectrum to discrete velocity values as the classification result \citep{park2020automatic, qiu2021attention}. Next, \cite{wang2021automatic} utilized U-Net as a regressor or a classifier, instead of a semantic segmentation tool as usual. Unlike the above classification or regression methods, \cite{zhang2019automatic} applied YOLO to detect the velocity points and utilized Long Short-Term Memory (LSTM) to correct the picked points. Later on, \cite{2021Ma} proposed a new network for object detection to improve picking accuracy. Additionally, \cite{wang2022automatic} proposed a fusion method of the spectrum and other information based on the semantic segmentation method to resolve auto-picking in the case of low Signal Noise Ratio (SNR). Among the above supervised learning-based approaches, deep learning models can achieve high picking accuracy with sufficient training data. However, in decreasing the number of training samples, supervised models perform negatively without the addition of other knowledge. Additionally, even though transfer learning has been introduced to alleviate the generalization problem \citep{park2020automatic}, the picking results are unreliable when the distribution of the work area changes significantly.

The unsupervised methods fuse more geophysical knowledge into models, improving the interpretive ability for automatic picking processing. They can be split into two categories: combinatorial optimization-based methods and clustering-based methods. The former usually selects a few reasonable points as a candidate point set (e.g., the local maximum of coherence values), and then utilizes some combinatorial optimization methods, e.g. feed-forward neural network, genetic algorithm, etc, to get the optimal combination as the final picking results \citep{Schmidt1992stacking, Fish1994NN, huang2013genetic, huang2016seismic}. Generally, researchers design objective functions based on the constraints of interval velocity or the slope of the velocity curve. However, the objective functions are chosen empirically and are hard to unify, which means that objective functions will change under different geological conditions. Unlike combinatorial optimization-based methods, clustering-based methods pay more attention to blob assumptions based on geophysical knowledge to split each energy peak. In general, they most often first split the points whose coherence values are larger than a fixed threshold, then apply clustering methods to obtain the centers of each blob as the picked points \citep{bin2019machine}. Compared to optimization-based methods, they have a less subjective variable selection, except for hyper-parameters, e.g., the number of clusters in K-means, the number of components in GMM (Gaussian mixture of models) \citep{bishop2006pattern}, etc. However, the above clustering methods only focus on the information of spatial location in the spectrum, but ignore the information on coherence values, greatly restricting the upper limit of picking accuracy. 
According to the description of the above-related works, we summarize the following problems that current methods have to face: (1) supervised learning-based methods rely heavily on training samples, and have poor generalizability across domains. (2) there are too many subjective choices in combinatorial optimization-based methods. (3) clustering-based methods lack the use of coherence values and local information of near spectra. 

Based on the above summary, there is a trade-off between dataset volume and picking accuracy in supervised methods. Thus, to alleviate dependence on data, we choose an unsupervised framework to resolve the picking problem. Additionally, we hope to further improve the accuracy of current unsupervised learning-based methods and maintain comparable performance to supervised methods. Therefore, we develop an unsupervised learning approach named Unsupervised Ensemble Learning (UEL) method, which fuses more information into the single-sample velocity spectrum. UEL consists of three main methods: a new spectrum gain method, a new clustering method, and a new ensemble learning method. 
First, we propose a gain method for the velocity spectrum to equalize the global signal as a preprocessing method. Though there is some work on the scanning calculation process of the velocity spectrum \citep{fomel2009velocity, luo2012velocity, chen2015velocity}, our proposed gain method can be applied directly to the velocity spectrum to further improve picking accuracy.
Second, to pick suitable blob centers from a velocity spectrum, we also propose the Attention Scale-Space Filter (ASSF) clustering method. Compared with the base model SSF \citep{leung2000clustering}, ASSF makes better use of the coherence information and pays more attention to the points with large coherence values. 
Finally, our UEL fuses the information of the current spectrum, the near spectra, and a few labeled spectra to solve the velocity analysis problem. The use of additional information can guide the current spectrum to filter out outliers. In a word, ensemble learning can improve the robustness of picking results very well, and even on datasets with median and low SNRs, UEL still produces stable picking results.

\section{METHODOLOGY}
\begin{figure}[bt]
\centering
\includegraphics[width=0.6\textwidth]{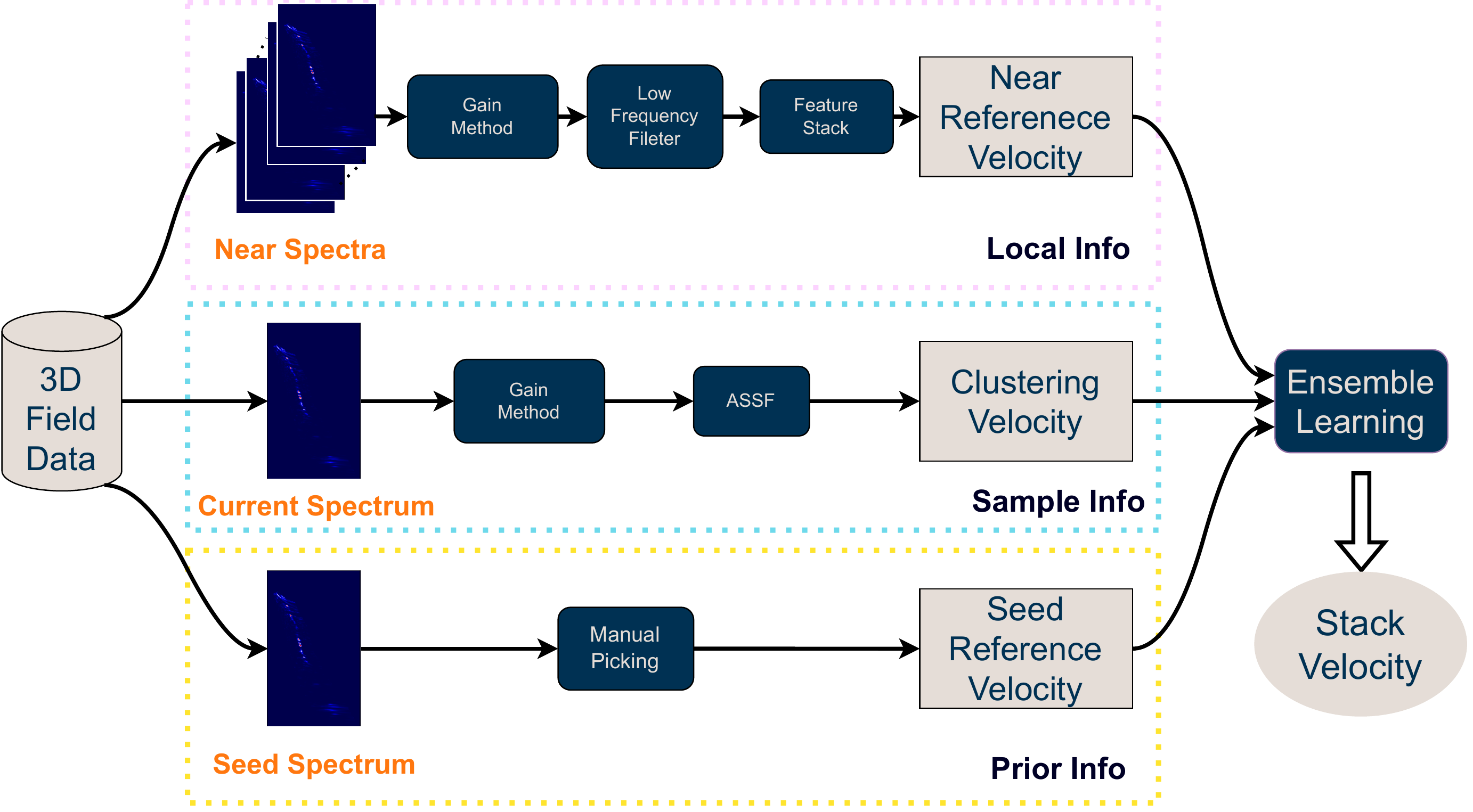}
\caption{The workflow of proposed UEL. We first perform the information gain method on the current velocity spectrum and utilize the ASSF method to cluster scatters of the gained spectrum to extract the sample information. Then, we define the near velocity spectra according to the location information and obtain local velocity information by sequentially implementing spectral gain, low-frequency feature filtering, and feature stacking. Next, the prior velocity information is extracted by referring to the manual picking of the nearest seed spectrum. Finally, we fuse three parts of velocity information to estimate the stack velocity of the current spectrum using an ensemble learning method.}
\label{fig:MainFlow}
\end{figure}

We propose an unsupervised learning method named UEL for picking stack velocity, where there are three parallel modules in UEL that can run simultaneously to extract sample, local, and prior velocity information, respectively, as shown in Figure~\ref{fig:MainFlow}. 
In this section, we will introduce the key techniques of UEL in order.

\subsection{Velocity Spectrum and Gain Method}
\begin{figure}[bt]
	\centering
	\subfloat[]{\includegraphics[height=2in]{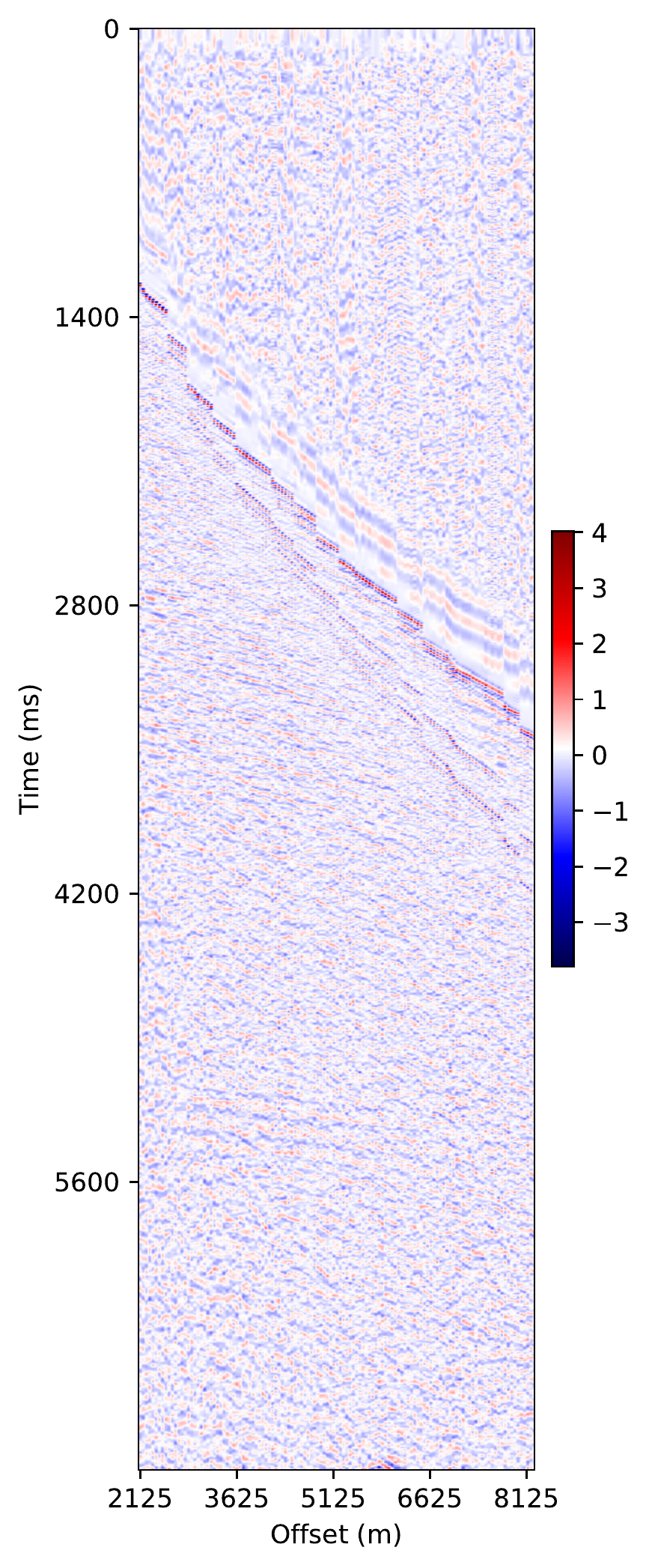}\label{fig:gain-OriGth}}
	\hfil
	\subfloat[]{\includegraphics[height=2in]{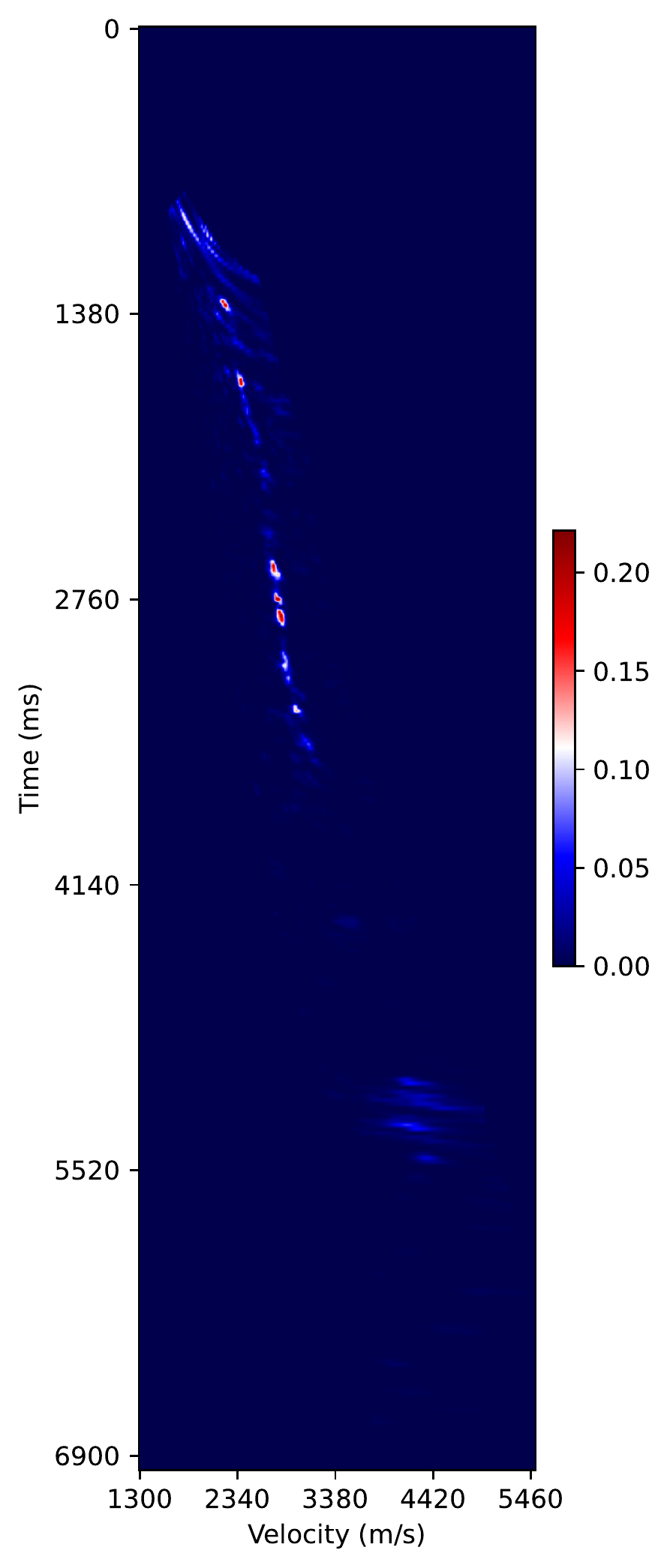}\label{fig:gain-Ori}}
	\hfil
	\subfloat[]{\includegraphics[height=2in]{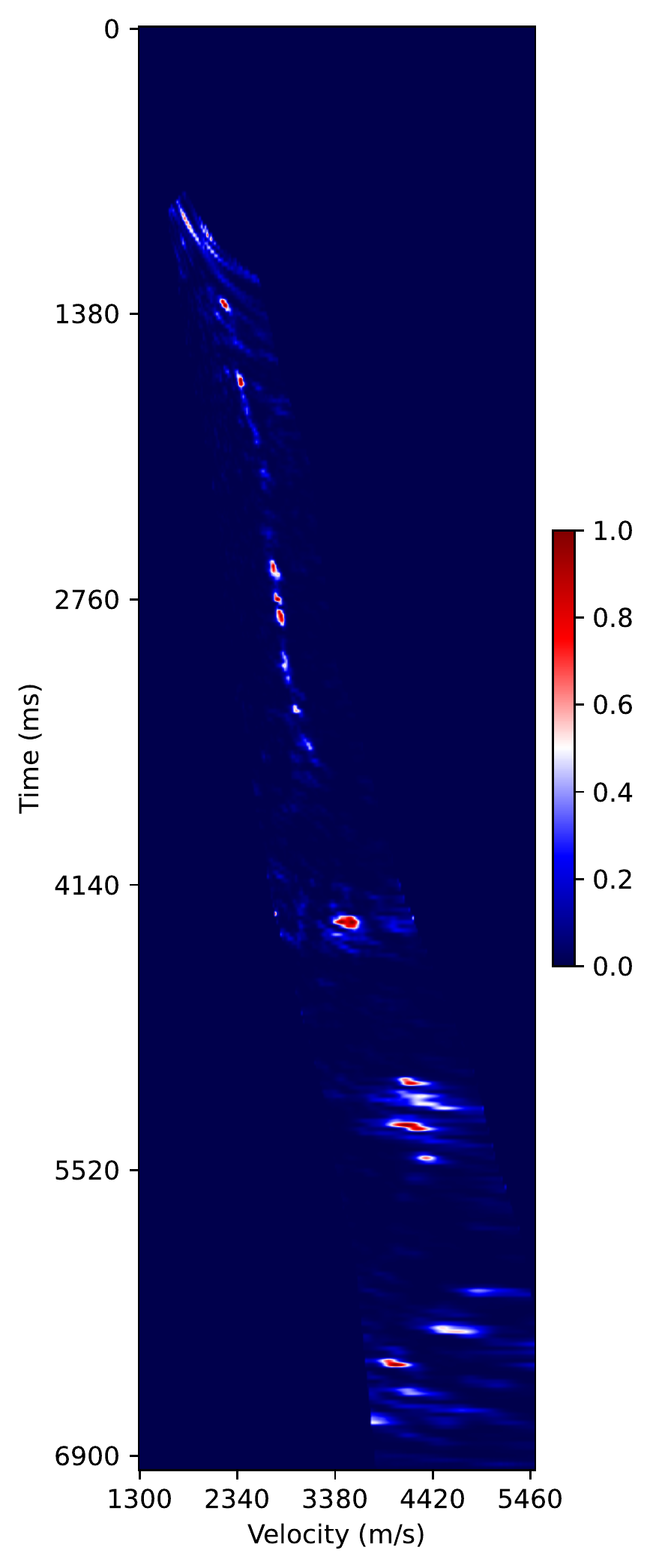}\label{fig:gain-LM}}
	\hfil
	\caption{The gain processing of the velocity spectrum: (a) original CMP gather; (b) original semblance velocity spectrum; (c) the gain result of (b).}
	\label{fig:gain}
\end{figure}

The velocity spectrum is generated by performing NMO correction on the CMP gathers (Figure~\ref{fig:gain-OriGth}) through different constant velocities, and then using some criteria to evaluate the coherence \citep{taner1969velocity}.
In this paper, we choose the semblance method \citep{neidell1971semblance} to compute the coherence, which is defined as:
\begin{equation}
    s[i]=\frac{\sum_{j=i-M}^{i+M}\left(\sum_{k=1}^{N} a[j, k]\right)^{2}}{N \sum_{j=i-M}^{i+M} \sum_{k=1}^{N} a[j, k]^{2}},
    \label{semblance}
\end{equation}
where $s[i]$ is the coherence value of $i$th time sample, $a[j, k]$ indicates the amplitude value at time index $j$ and trace number $k$ of the NMO gather. $N$ is the number of all traces, and $M$ is the summation range of the time axis. In this case, $2M+1$ is the summation length of the local window. Eq. \ref{semblance} computes the coherence values of an NMO gather at different times, and the velocity spectrum is calculated from multiple NMO gathers scanned by multiple NMO velocities. Therefore, 
we can concatenate these coherence vectors to form the value matrix of the velocity spectrum.
This method measures the correlation between the parts of the NMO gather and the constants to measure the coherence.

As Figure~\ref{fig:gain-Ori} shows, the signal strength of the reflection event is different, which leads to the inconsistency of the coherence scale of these energy clusters in the seismic velocity spectrum. To balance the scale, we propose a velocity spectrum gain method using Local Normalization (LN). LN method divides each element $C_{i,j}$ in the matrix by the local mean, that is, 
\begin{equation}
C_{i, j}^* = \left\{
\begin{array}{lcl}
\frac{C_{i, j} \cdot (i+L)}{ {\textstyle \sum_{k = 1}^{i+L} C_{k,j}}},  & & {0 < i \le L;}\\ 
\frac{C_{i, j} \cdot (1+2L)}{ {\textstyle \sum_{k = i-L}^{i+L} C_{k,j}} }, & & {L < i \le H-L;}\\
\frac{C_{i, j} \cdot (H+L-i+1)}{ {\textstyle \sum_{k = i-L}^{H} C_{k,j}} }, & & {H-L < i \le H,}
\end{array} \right.
\label{LN}
\end{equation}
where $C_{i, j}^*$ and $C_{i, j}$ are the values at the $i$th time sample and the $j$th velocity in the gained and original spectra, respectively. $L$ is the mean range, and $H$ is the maximum value of time index.

As Figure~\ref{fig:gain-LM} shows, our gain method balances the global scale very well, and the values of coherence are mapped to the range of $[0, 1]$. The energy blobs of the shallow and deep layers are easier to be observed. Not only that, our gain method does not change the location of the local extreme, which ensures the accuracy of the later automatic picking algorithm. 

\subsection{Near Reference Velocity}
In 3D-field seismic data, the geological layers are continuous in space. Thus, the near velocity spectra can also provide the current velocity spectrum with some velocity prior information. The reference velocity picking method of the near spectra is composed of four main parts. 

\begin{figure}[bt]
	\centering
	\subfloat[]{\includegraphics[height=3in]{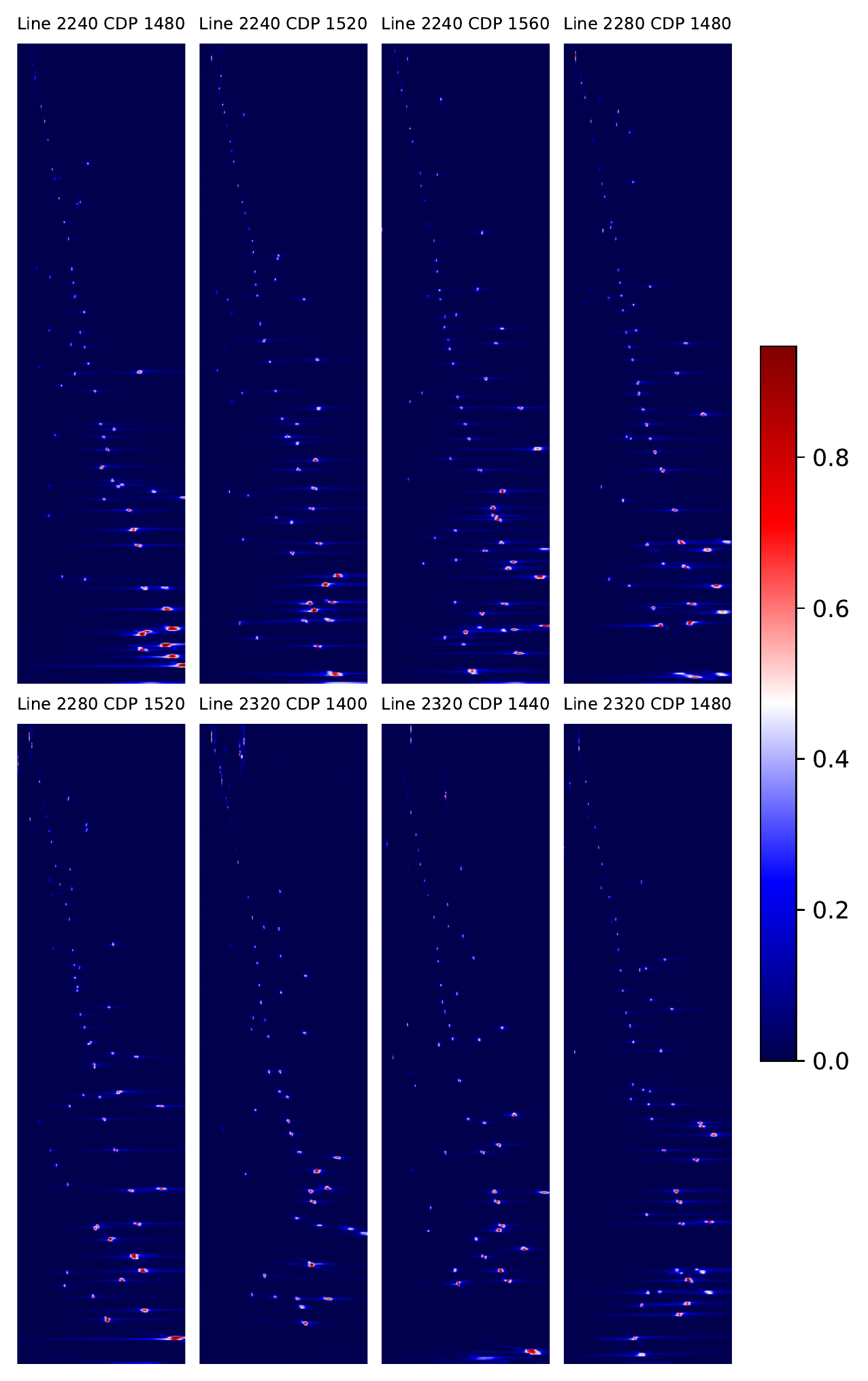}\label{fig:NearPwr}}
	\hfil
	\subfloat[]{\includegraphics[height=3in]{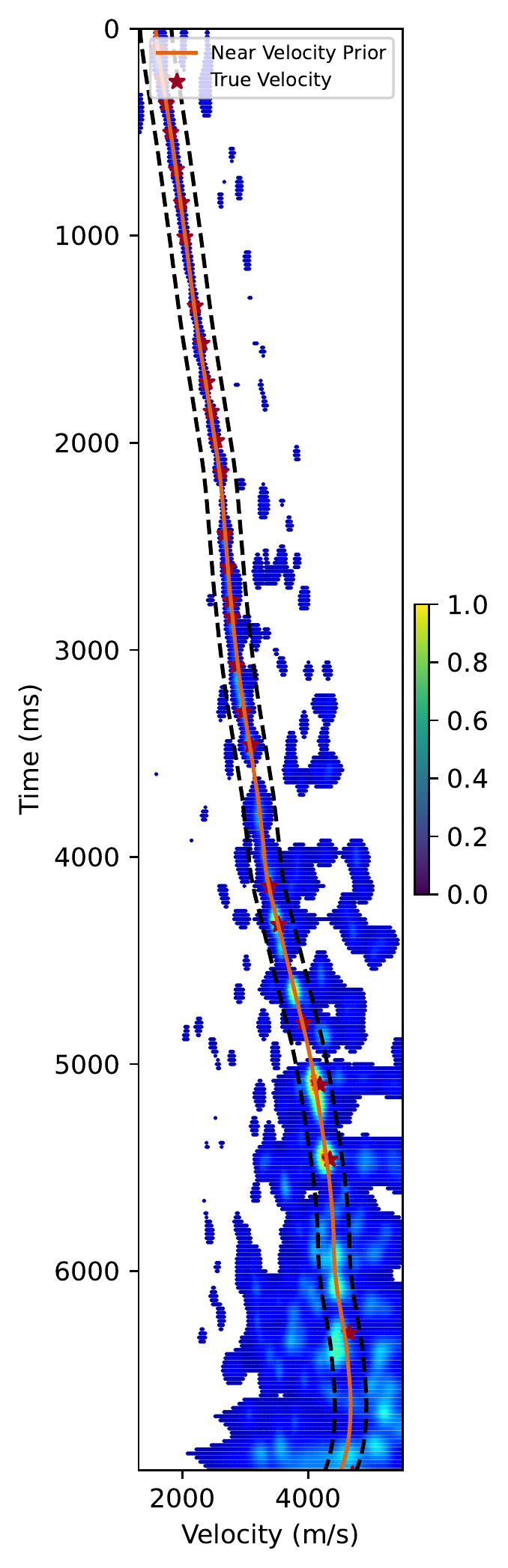}\label{fig:NearPwrRef}}
	\caption{(a) The gained spectrum images of near spectra. (b) The stacked low-frequency map and the visual result of ALWLR. In (b), the red stars are the true stack velocity points, the black dotted curves are the confidence boundaries, and the orange curve is the estimated near reference velocity curve.}
	\label{fig:NearVelocity}
\end{figure}

First, we define the velocity spectra satisfying the following two situations as the near velocity spectra:
\begin{itemize}
    \item The same line number and near CDP numbers;
    \item The same CDP number and near line numbers.
\end{itemize}
Then, the near spectra also are gained by the above LN gain method and are shown in Figure~\ref{fig:NearPwr}.

Second, we use a low-pass filter to obtain the low-frequency information of each near spectrum. In this paper, we choose the average blur to filter spectra, and the kernel matrix is defined as:
\begin{equation}
    {\left[ {\begin{array}{*{20}{c}}
w^{-2}& \cdots &w^{-2}\\
 \vdots &w^{-2}& \vdots \\
w^{-2}& \cdots &w^{-2}
\end{array}} \right]_{w \times w}},
\label{blur}
\end{equation}
where $w$ is the width of the kernel.

Next, these low-frequency feature maps of the near spectra are stacked to get common information by
\begin{equation}
    S_{i,j}^* = \sum\limits_{k = 1}^K {S_{i,j}^{\left( k \right)}/K},
    \label{SumLF}
\end{equation}
where $S_{i,j}^*$ is the element of the stacked feature map, $S_{i,j}^{(k)}$ is the element of the $k$th low-frequency feature map, and $K$ is the number of all near spectra. In Figure~\ref{fig:NearPwrRef}, before estimating the velocity curve, the velocity spectrum is transformed into a three-dimension point set $\left\{ {\left( {{t_i},{v_i},{c_i}} \right)} \right\}_{i = 1}^n$ by using a splitting threshold, and $c_i$ is the coherence value of point $i$.

Finally, the reasonable reference velocity is estimated by the Attention Locally Weighted Linear Regression (ALWLR). Our proposed ALWLR is an improved algorithm of Locally Weighted Linear Regression (LWLR), which is a non-parametric algorithm, and is always used to solve the underfitting problem \citep{hardle2004nonparametric}. In the local regression of LWLR at time $t_0$, the loss function is given as:
\begin{equation}
    loss(t_0) = {\sum\nolimits_i {{w_i}\left( {{v_i} - {t_i} \cdot \alpha  - \beta } \right)} ^2},
    \label{LWLR-loss}
\end{equation}
with
\begin{equation}
    {w_i} = {e^{ - \frac{{{{\left| {{t_i} - {t_0}} \right|}^2}}}{{2h}}}},
    \label{LWLR-weight}
\end{equation}
where $\alpha, \beta$ are the regression parameters. $t_i$, $v_i$ are the time sampling value and velocity at point $i$, respectively, and $h$ is the bandwidth of the Gaussian kernel.
LWLR uses the local sample location information at the prediction point $t_0$ to construct a weighted kernel (Eq. \ref{LWLR-weight}). While in ALWLR, we purpose the following weighted kernel so that more attention is paid to the coherence value of each point, i.e.,
\begin{equation}
    {w_i} = c_i^\lambda {e^{ - \frac{{{{\left| {{t_i} - {t_0}} \right|}^2}}}{{2h}}}},
    \label{ALWLR-weight}
\end{equation}
where $c_i$ is the coherence value at point $i$ and $\lambda$ is the weight hyper-parameter between two parts. In this paper, we choose $\lambda = 5$. The implementations of ALWLR are shown in Algorithm~1.

In Figure~\ref{fig:NearPwrRef}, the velocity estimation of ALWLR considers both the location information as well as the coherence information, and the true stack velocity points are all in the neighborhood of the curve. This velocity estimation provides the following ensemble method with good velocity prior information. 

\begin{algorithm}[ht]
    \caption{ALWLR Algorithm}
    \begin{algorithmic}\label{ALG:ALWLR}
    \State \textbf{Input:} samples $\{(t_i, v_i, c_i)\}^n_{i=1}$, time sampling set $\{t_j^*\}^m_{j=1}$, bandwidth of Gaussian kernel $h$, coherence weight $\lambda$
    \newline
    \For{ j = 1,..., m}
	\State \textit{1.} ${w_i} = c_i^\lambda {e^{ - \frac{{{{\left| {{t_i} - {t_j^*}} \right|}^2}}}{{2h}}}}, i=1,\dots, n, \boldsymbol{W_j} = \text{diag}(w_{1},...,w_{n)})$;
        \State \textit{2.} $\boldsymbol{X_j} = {\left[ {\begin{array}{*{20}{c}}{{t_1}}& \cdots &{{t_n}}\\1& \cdots &1\end{array}} \right]^T}, \boldsymbol{Y_j} = {\left[ {{v_1}, \cdots ,{v_n}} \right]^T}, x_j^* = \left[ {\begin{array}{*{20}{c}}{t_j^*}&1\end{array}} \right]$;
        \State \textit{3.} $v_j^* = x_j^* \cdot {\left( {\boldsymbol{X_j}^T{\boldsymbol{W_j}}{\boldsymbol{X_j}}} \right)^{ - 1}}\boldsymbol{X_j}^T{\boldsymbol{W_j}}{\boldsymbol{Y_j}}$;
    \EndFor
	
    \State \textbf{Output: } predicted velocity: $\{v_j^*\}^m_{j=1}$
    \end{algorithmic} 
\end{algorithm}

\subsection{Cluster by Attention Scale-Space Filter}
When manually picking up the velocity spectrum, the analyst generally finds the energy centers to be selected first. We hope to imitate the process by which the human eye perceives things to select energy centers. Scale-Space Filter (SSF) clustering method is a clustering method based on scale space theory and can find suitable cluster centers in a scale space \citep{leung2000clustering}. Thus, we select the SSF method as the base model of our clustering method. 
Since the points in the velocity spectrum have additional attribute values, we propose an Attention SSF (ASSF) clustering method in which the coherence information is included.

In the scale space theory, there are two important concepts. We take a two-dimensional image as an example to interpret. One is the initial Probability Density Function (PDF) of the samples $p(x):\mathrm{R^2}  \mapsto \mathrm{R}$. The other is the PDF of the samples at a scale $P(x, \sigma)$, and $\sigma$ is the scale parameter. 
The scale of observation describes the change in the focal length of the human eye when a person observes an object. For example, when looking closely at close range, $\sigma$ is 0, and when the human eye is looking away from the object, things become blurred and $\sigma$ starts to get larger.
$P(x, \sigma)$ is usually represented by the convolution of $p(x)$ with Gaussian kernel:
\begin{equation}
    P(x, \sigma)=p(x) * g(x, \sigma)=\int p(x-y) \frac{1}{\left(\sigma^{2} 2 \pi\right)} e^{-\frac{\|y\|^{2}}{2 \sigma^{2}}} d y,
    \label{SSF}
\end{equation}
where $\sigma$ is the scale parameter, and $g(x, \sigma)$ is the Gaussian kernel:
\begin{equation}
    g(x, \sigma)=\frac{1}{(\sigma \sqrt{2 \pi})^{2}} e^{-\frac{\|x\|^{2}}{2 \sigma^{2}}}.
    \label{GKer}
\end{equation}
Based on the scale space theory, SSF derives the class-center iterative convergence formula of the $(n+1)$th iteration at a specific scale $\sigma$:
\begin{equation}
    x(n+1)=\frac{\sum_{i=1}^{N} x_{i} e^{-\frac{\left\|x(n)-x_{i}\right\|^{2}}{2 \sigma^{2}}}}{\sum_{i=1}^{N} e^{-\frac{\left\|x(n)-x_{i}\right\|^{2}}{2 \sigma^{2}}}},
    \label{SSF-update}
\end{equation}
where $n$ denotes the iteration number. Eq. \ref{SSF-update} is a weighted shift processing, and the shift weight of $x_i$ is based on the Gaussian kernel of the Euclidean distance between $x_i$ and $x(n)$. When the center points converge (Eq. \ref{SSF-update}), the nearer center points are merged, and the merging result is used as the cluster center under the scale $\sigma$. The change rate of the scale parameter $\sigma$ is consistent with the blurring process observed by the human eye, which is 0.029 \citep{leung2000clustering}. 

In ASSF, we pay more attention to the coherence information and add it to the model assumption. We assume that 
$$p\left( {{x_i}} \right) \propto {c_i},$$ and $c_i$ is the coherence value of $x_i$. In the $(n+1)$th iteration, the class centers are updated as:
\begin{equation}
    x(n + 1) = \frac{{\sum\limits_{i = 1}^N {{x_i}c_i^\alpha } {e^{ - \frac{{{{\left\| {x(n) - {x_i}} \right\|}^2}}}{{2{\sigma ^2}}}}}}}{{\sum\limits_{i = 1}^N {c_i^\alpha {e^{ - \frac{{{{\left\| {x(n) - {x_i}} \right\|}^2}}}{{2{\sigma ^2}}}}}} }},
    \label{ASSF-update}
\end{equation}
where $n$ denotes the iteration number, $\alpha$ is the weight parameter between the Gaussian kernel and coherence kernel, and we set $\alpha = 1$ in this work. 
To confirm the suitable cluster number, we define the lifetime of each cluster number, and its concept is consistent with hierarchical clustering. In ASSF, the lifetime is the longest $\sigma$ iteration time when the cluster number does not change. 

In the specific algorithm implementation, we set the minimum number of centers to ensure the picking rate, and we suggest it should be set from 10 to 20. Moreover, the merging condition and the convergence condition of the centers are set according to geophysical knowledge. When the data is not standardized, the merging threshold is usually set between 100-200, and the convergence threshold is usually set between 10-50. In summary, the pseudocodes of the ASSF clustering algorithm are summarized in Algorithm~2.

\begin{algorithm}[bt]
    \caption{ASSF Clustering Algorithm}
    \begin{algorithmic}\label{ALG:ASSF}
    \State \textbf{Input:} samples $\{x_i(t_i, v_i), c_i\}^n_{i=1}$, initial scale parameter $\sigma_0$, bandwidth of Gaussian kernel $h$, center minimum number $K$, merging threshold $T_m$, convergence threshold $T_c$, coherence weight $\alpha$;
    \State \textbf{Initialize:} a) center set $C=\{c^{0}_i\}^n_{i=1}=\{x_i\}^n_{i=1}$, b) the set of center set $\mathscr{C}=\{C\}$, c) the scale parameter $\sigma = \sigma_0$; %
    \While{center number $\#(\mathscr{C})> K$}
        \State \textbf{1.} $k = 1, C_0 = C$;
        \State \While{$\mathop {\min }\limits_j \left\{ {\left\| {c_j^{k - 1} - c_j^k} \right\|_2^2} \right\} > {T_c}$}
        \State \textbf{2.} update the center point $c_j^{k + 1} = \frac{{\sum\limits_{i = 1}^n {{x_i}c_i^\alpha } {e^{ - \frac{{{{\left\| {c_j^k - {x_i}} \right\|}^2}}}{{2{\sigma ^2}}}}}}}{{\sum\limits_{i = 1}^n {c_i^\alpha {e^{ - \frac{{{{\left\| {c_j^k - {x_i}} \right\|}^2}}}{{2{\sigma ^2}}}}}} }}, N = k$
        \EndWhile
        \State \textbf{3.} merge the points in $C_N$ whose pairwise distance $< T_m$, and get new center set $C_N$;
        \State \textbf{4.} update the scale parameter $\sigma = 1.029\sigma$;
        \State \textbf{5.} log the initial value of next iteration $C = C_N$;
        \State \textbf{6.} add $C_N$ into $\mathscr{C}$
    \EndWhile
    \State \textbf{7.} Search the set $C_{final}$ which has maximum lifetime;
    \State \textbf{Output:} cluster center set $C_{final}$ 
    \end{algorithmic}
\end{algorithm}

\subsection{Ensemble Learning for Velocity Picking}
The clustering method can only detect some candidate points like RMS velocities. To improve the accuracy of automatic picking, we propose a new Unsupervised Ensemble Learning (UEL) method to fuse the information provided by the near spectra and the seed velocity with the picking results of ASSF. 

First, we define the seed velocity. In a 3-D field dataset, we sample evenly on the space at a certain sampling rate and deem these samples as seed points. The stack velocities of these seed point velocity spectra are picked up manually in advance. According to the continuity of the stratum, the manually picked velocity of the seed point closest to the current velocity spectrum can be taken as the reference velocity. 

Then, we construct a confidence area on the current spectrum based on the reference velocity curve of the near seed points $V_{RS}$ and the reference velocity curve of  near spectra $V_{RN}$. The velocity spectrum can be defined as a two-dimensional point set:
\begin{equation}
    {S = \left\{ {\left( {t,v} \right):t \in \left( {{T_{\min }},{T_{\max }}} \right),v \in \left( {{V_{\min }},{V_{\max }}} \right)} \right\}},
    \label{pwr}
\end{equation}
where $T_{\min }$,$T_{\max }$,$V_{\min }$, and $V_{\max }$ are the minimum and maximum of the time sample index and velocity sample index, respectively.
We set the confidence parameter $w$ to control the range of the confidence area, and use the following function to define the Confidence Area (CA) on the spectrum:
\begin{equation}
    \text{CA}=\{x: \text{D}(x, V_{RS}) < w, \text{D}(x, V_{RN}) < w, x \in S\},
    \label{eq:RefCons}
\end{equation}
where set $S$ is defined by Eq. \ref{pwr} and $\text{D}(x, c)$ is the Euclidean distance between the point $x$ and the curve $c$.
Next, the CA can guide the cluster centers obtained from ASSF $C_1$ to select reasonable centers. Specifically, we choose the center of the set $C_1$ that falls in the CA area and denote them as $C_2$.

Finally, we use interval velocity constraint to remove abnormal points of $C_2$. The interval velocity can be computed by the stack velocities using Dix Formula \citep{dix1955seismic}, as follows:
\begin{equation}\label{eq: Dix}
	V_i^{int} = {\left( {\frac{{V_n^2{t_n} - V_{n - 1}^2{t_{n - 1}}}}{{{t_n} - {t_{n - 1}}}}} \right)^{1/2}},
\end{equation}
where $V_n$, $t_n$ and $V_{n-1}$, $t_{n-1}$ are the stack velocity and the time at the $n$th and $(n-1)$th picking points, and $V_i^{int}$ is the interval velocity between $t_n$ and $t_{n-1}$. Actually, it is not reasonable to perform the interval velocity constraint for all auto-picking points. There is a minimum interval time between the auto-picking points. In our work, we set 200ms as the minimum interval time. When the time interval of two adjacent picking points is below 200ms, one point is removed. The implementation of the interval velocity constraint is actually a recursive algorithm, as shown in Algorithm 3.

\begin{algorithm}[bt]
    \caption{Interval Velocity Constraint in UEL}
    \begin{algorithmic}\label{ALG:IntervalVel}
    \State \textbf{Input:} picking result $C_p$, seed reference velocity $V_{RS}$, near spectra reference velocity $V_{RN}$, the range of interval velocity $(V_{min}, V_{max})$, minimum time interval $T_M$;
    \State Initialize $C_I = C_p$, and sort $C_I=\{C_i^I(t_i, v_i)\}_{i=1}^M$ by time domain;
    \While{True}
        \State $N = \#(C_I), k = 0$ \quad// $\#(\cdot)$: count element number
        \For{$i=1:(N-1)$}
            \If{$t_{i+1}-t_{i} < T_M$}
                \State Remove the point $C_j^I$ far from both $V_{RS}$, $V_{RN}$;
                \State $C_I = C_I - \{C_j^I\}$, $k = 1$;
                \State \textbf{break} \quad// break for
            \EndIf
            \State $\text{IntV} = \text{IntervalVel}(C_i^E, C_{i+1}^E)$ \quad// interval velocity constraint
            \If{$\text{IntV} \notin (V_{min}, V_{max})$}
                \State Remove the point $C_j^I$ far from both $V_{RS}$, $V_{RN}$;
                \State $C_I = C_I - \{C_j^I\}$, $k = 1$;
                \State \textbf{break} \quad// break for
            \EndIf
        \EndFor
            
        \If{$k = 0$}
           \State $C_E = C_I$;
           \State \textbf{break} \quad// break while
        \EndIf
    \EndWhile
    \State \textbf{Output:} Ensemble picking set $C_E$
    \end{algorithmic}
\end{algorithm}

\section{EXPERIMENTS}
\subsection{Datasets}
To evaluate our method, we choose both synthetic datasets and a field dataset. The field dataset is a 3-D field dataset with medium SNR in China, denoted as dataset A. In dataset A, there are 4453 spectra, of which the stacking velocity of 1085 spectra is picked manually. The sampling interval of the time dimension and velocity dimension are 20ms and 20 m/s, respectively. The sampling ranges of these two dimensions are $[0, 6960]$ and $[1300, 5500]$, respectively. 
The synthetic datasets are generated by the known stack velocity field. We add Gaussian noise to the known velocity field to simulate the noise in the field data and then define the SNR as
\begin{equation}
    SNR = \frac{\# \left( {{V_{real}}} \right)}{\# \left( {{V_{noise}}} \right)},
    \label{SNR}
\end{equation}
where $\# \left( {{V_{real}}} \right)$ and $\# \left( {{V_{noise}}} \right)$ are the numbers of stack velocity points of the real model and added Gaussian noise, respectively. These stack velocity points are used to compute the prestack gathers and then calculate the velocity spectra. In this paper, we choose the stack velocity field as the base model, which is used to generate synthetic datasets. We generate eight synthetic datasets with SNRs of 10, 4, 2, 1, 2/3, 1/2, 2/5, 1/3, and denoted as S1 - S8, respectively.

\subsection{Quality Control}
To evaluate the accuracy and robustness of picked stack velocities, we perform the quality control both quantitatively and qualitatively. Quantitatively, we define Velocity Mean Absolute Error (VMAE) and Velocity Mean Relative Error (VMRE) to measure the absolute error and relative error between two discrete stack velocity curves, respectively, that is, 
\begin{equation}
    \text{VMAE}\left( {{V}_A,{V}_M} \right) = \frac{1}{N}\sum\limits_{i = 1}^N {\left| {V_i^m - V_i^a} \right|},
    \label{VMAE}
\end{equation}
\begin{equation}
    \text{VMRE}\left( {{V}_A,{V}_M} \right) = \frac{1}{N}\sum\limits_{i = 1}^N {\left| {V_i^m - V_i^a} \right|/V_i^m},
    \label{VMRE}
\end{equation}
where $V_A$ and $V_M$ are the point sets generated by automatic and manual picking methods, respectively. $N$ is the number of sampling points on the velocity curve. Concretely, $V_i^a$ and $V_i^m$ are the velocities at the $i$th time sample, respectively.

To better evaluate the performance of the synthetic seismic data, we define the Picking Rate (PR) and the Mean Deviation (MD) of the real stack velocity points. We assume that a real stack velocity point is correctly recognized if there exists a picking point in its neighborhood. 
Noting that real stack velocity points $\{(t_i^R,v_i^R)\}_{i=1}^N$ are known, we define PR as
\begin{equation}
    \text{PR} = {\frac{1}{N}} \cdot \sum\limits_{j = 1}^N {I\left\{ {\left| {v_i^A - v_i^R} \right| < 200} \right\}},
    \label{PR}
\end{equation}
where $I\{\cdot\}$ is the indicator function, $N$ is the number of real RMS velocity points, $v_i^A$ is the velocity which is picked by an algorithm at $t_i^R$.
After calculating PR, the recognized real RMS velocity points have been identified and recorded as $\left\{ {\left( {t_j^{R'}, v_j^{R'}} \right)} \right\}_{j = 1}^M$ and $M$ is the number of the recognized real RMS velocity points.
Next, we also measure the average deviation of picking among the recognized real RMS velocity points. Thus, we define the MD as 
\begin{equation}
    \text{MD} = {\frac{1}{M}} \cdot \sum\limits_{i = 1}^M {\left| {v_j^A - v_j^{R'}} \right|},
    \label{MD}
\end{equation}
where $v_j^A$ represents the velocity of the automatic picking curve at time $t_j^{R'}$, which corresponds to real RMS velocity $v_j^{R'}$.

Qualitatively, to measure the rationality of picked stack velocities, we use the flatness of the NMO gather which is corrected by the stack velocities, and the clarity of layer structure on the stack section of NMO gathers. 

\subsection{Performance of UEL}
\begin{figure}[bt]
    \centering
    \subfloat[]{\includegraphics[height=2in]{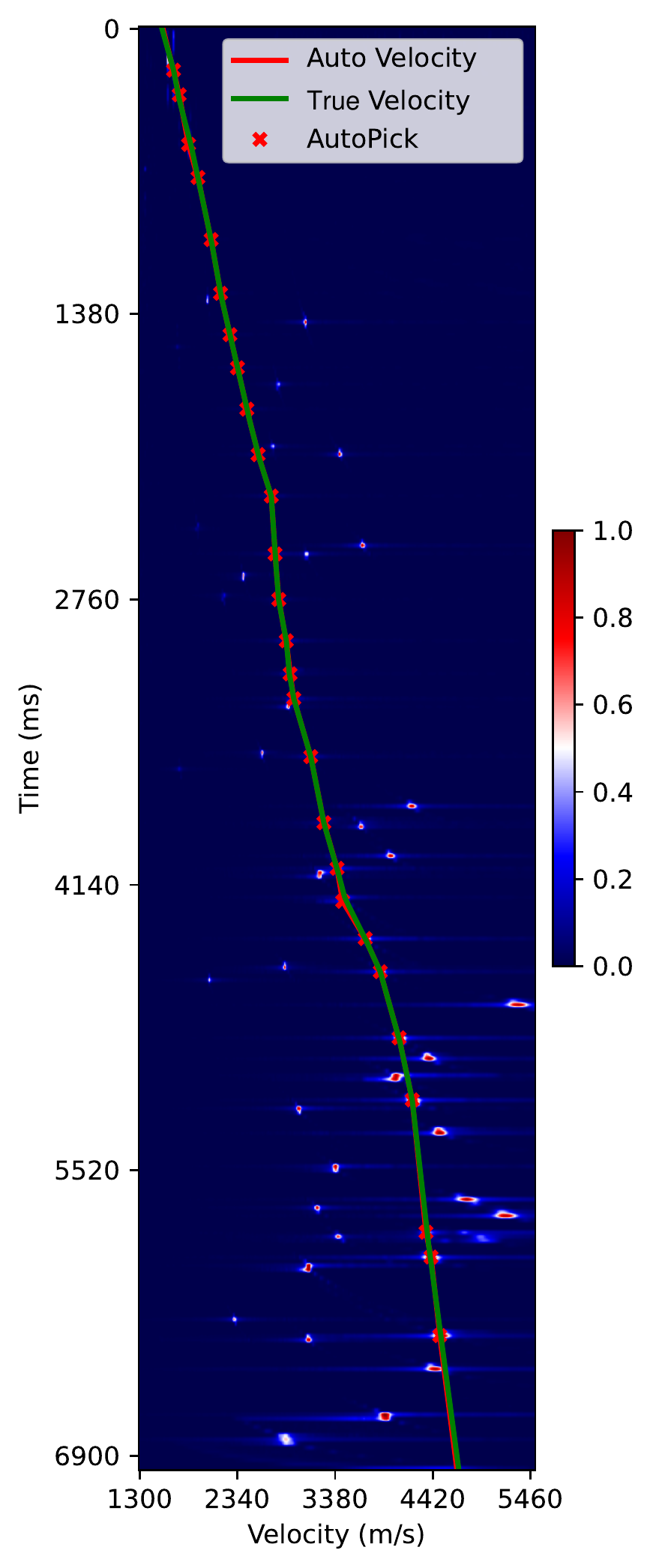}\label{fig:PWRwMPick-1}}
    \hfil
    \subfloat[]{\includegraphics[height=2in]{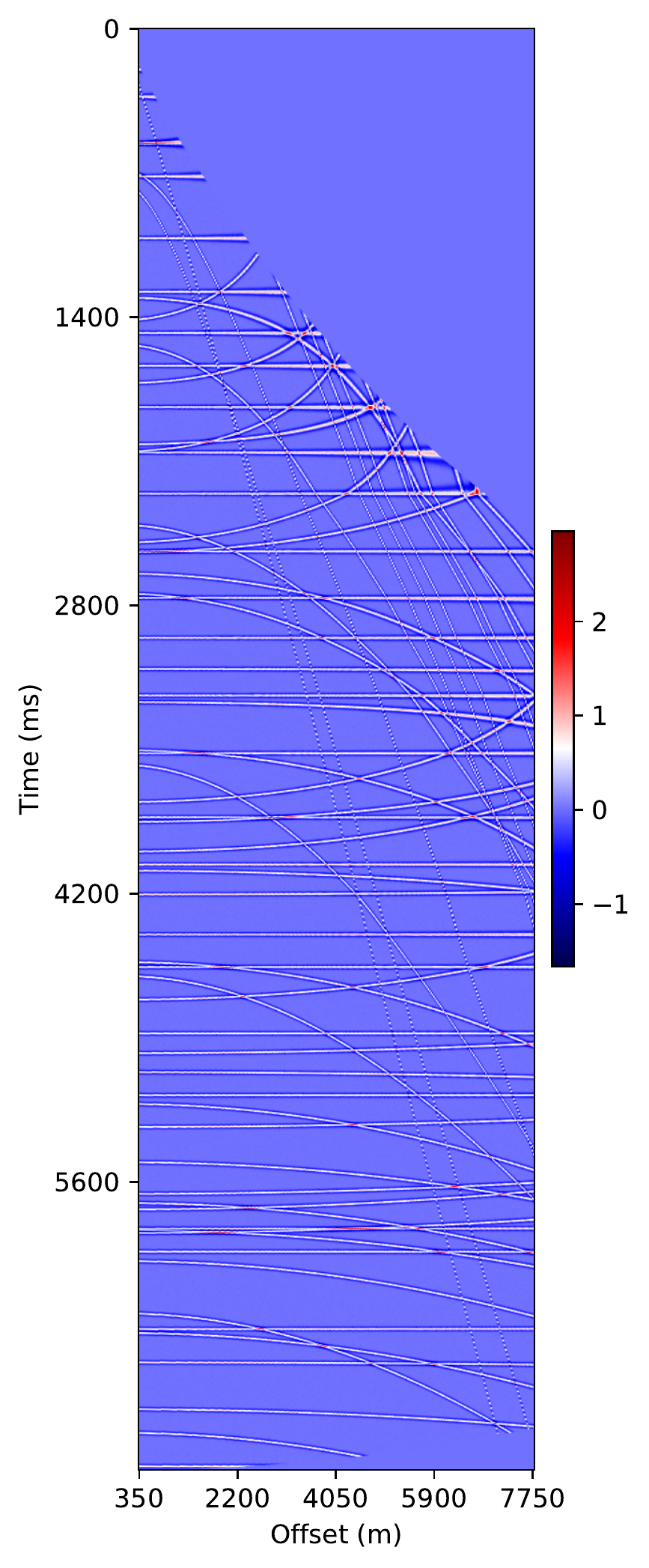}\label{fig:NMOGth-1}}
    \hfil
    \subfloat[]{\includegraphics[height=2in]{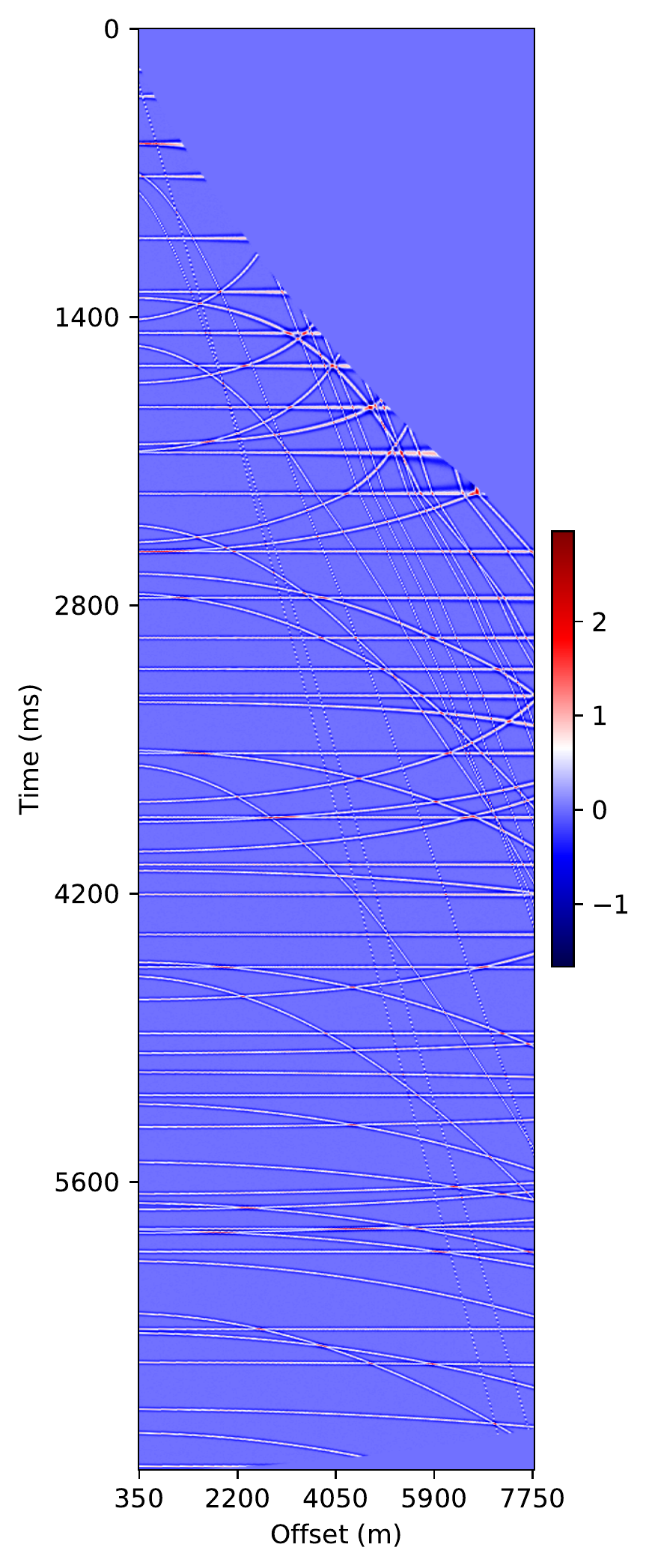}\label{fig:NMOGthGT-1}}
    \\
    \subfloat[]{\includegraphics[height=2in]{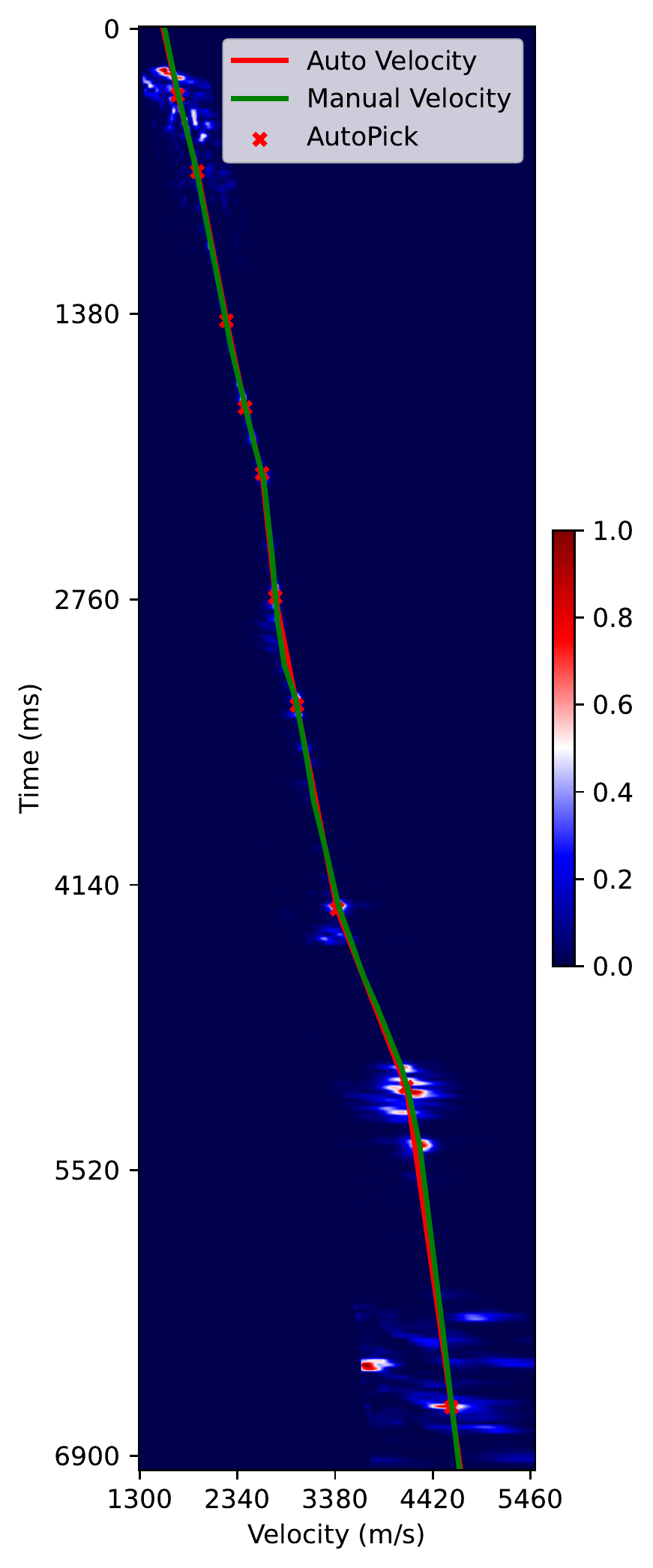}\label{fig:PWRwMPick-2}}
    \hfil
    \subfloat[]{\includegraphics[height=2in]{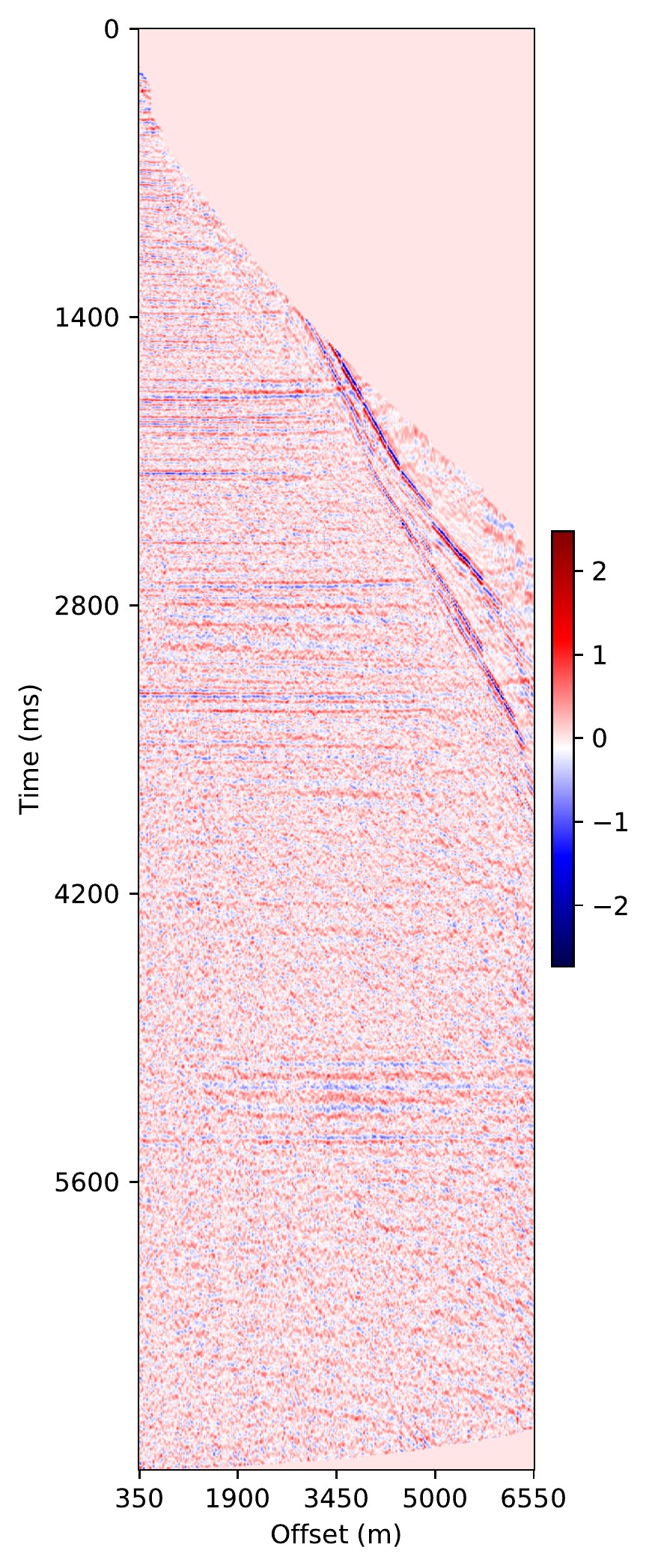}\label{fig:NMOGth-2}}
    \hfil
    \subfloat[]{\includegraphics[height=2in]{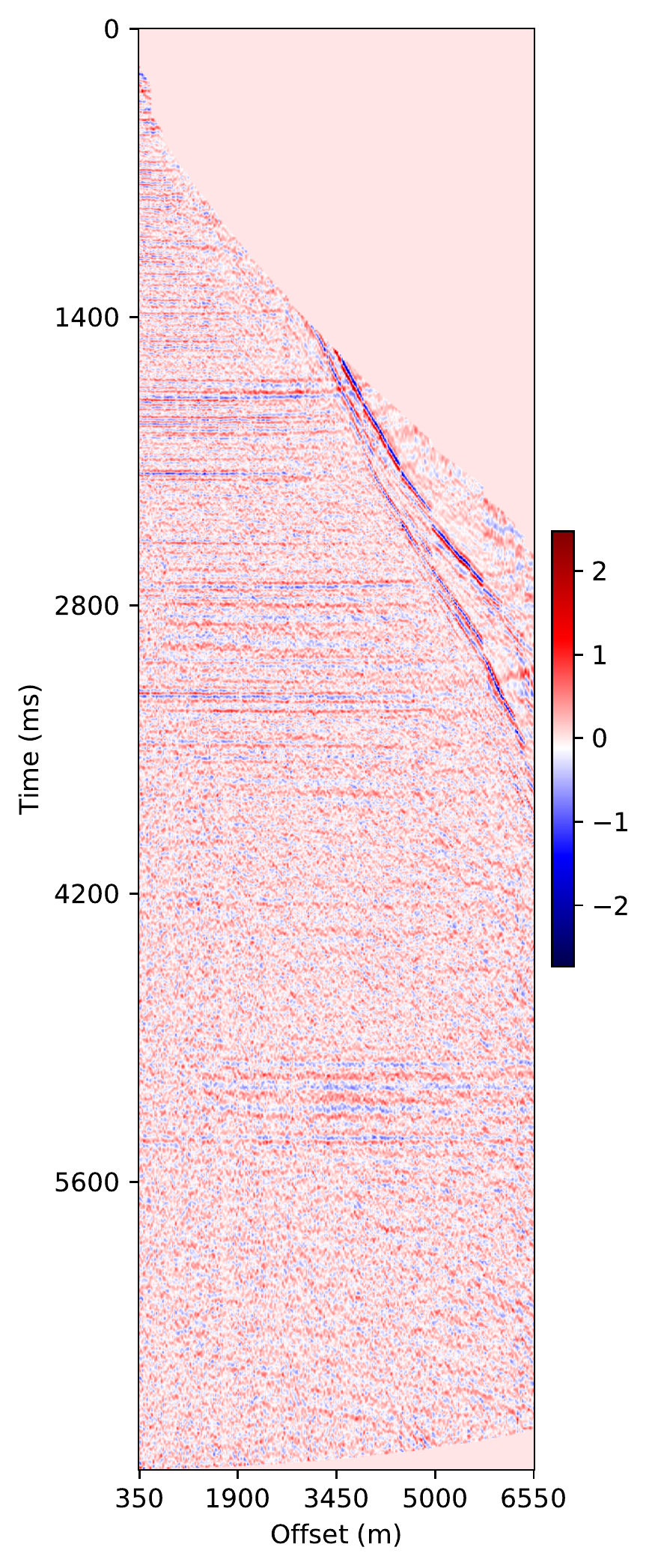}\label{fig:NMOGthGT-2}}
    \hfil
    \caption{The picking results of UEL. (a) and (c) are the gained spectrum with the velocity curves of UEL (red) and label (green) of S5 and A, respectively. (b), (c) and (e), (f) are the NMO correction gather based on UEL picking and label picking of S5 and A, respectively.}
    \label{fig:UEL-perf}
\end{figure}

\begin{figure}[bt]
    \centering
    \subfloat[]{\includegraphics[height=2in]{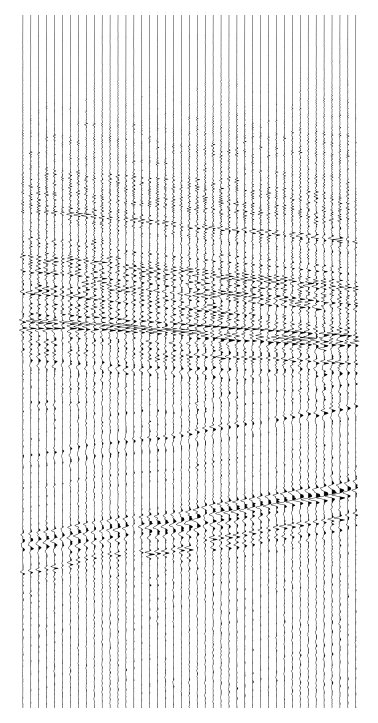}\label{fig:stk-2560-A}}
    \hfil
    \subfloat[]{\includegraphics[height=2in]{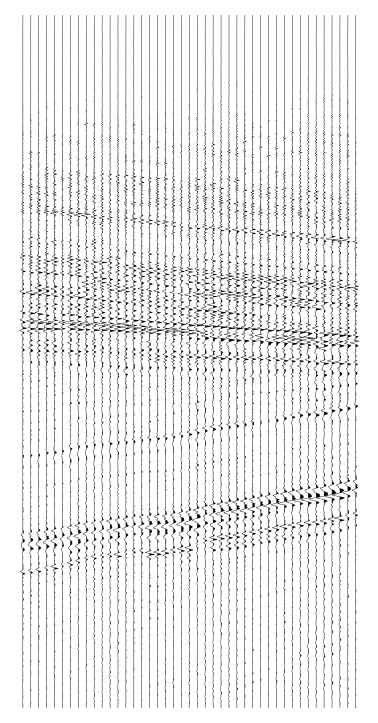}\label{fig:stk-2560-M}}
    \hfil
    \subfloat[]{\includegraphics[height=2in]{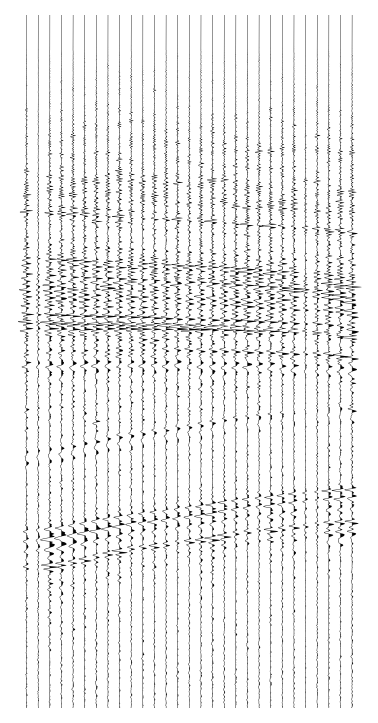}\label{fig:stk-2840-A}}
    \hfil
    \subfloat[]{\includegraphics[height=2in]{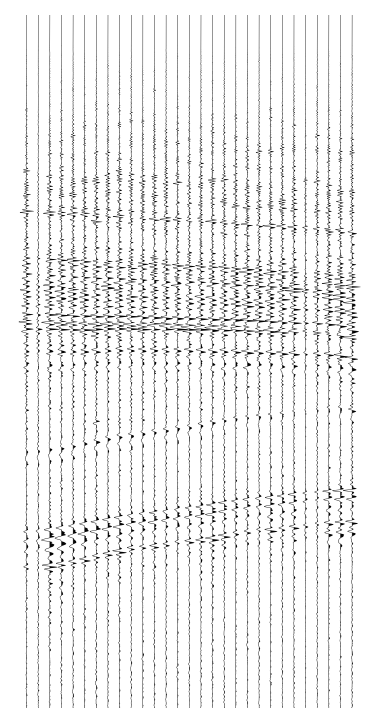}\label{fig:stk-2840-M}}
    \caption{The stack sections of NMO CMP gathers on dataset A are shown in (g)-(j). (g) and (h) are automatic and manual results of line-2560, respectively. (i) and (j) are automatic and manual results of line-2840, respectively.}
    \label{fig:stk-perf}
\end{figure}

To test the picking ability of our UEL, we conduct quantitative analysis and visual analysis on both synthetic and field datasets. First, we conduct quantitative tests on synthetic datasets with high, medium, and low SNRs, named S1, S3, S5, and a field dataset with medium SNR, named A, and the test results are shown in Table~\ref{tab:UELTest}. Our UEL achieves VMAE of 9.875m/s and 18.227m/s on synthetic datasets with medium and high SNRs, respectively, and both VMREs are less than 1\%. Even on the low SNR synthetic dataset S5, the VMAE is 45.274m/s and VMRE is 1.441\%, which can play a guiding role in the application to a certain extent. Moreover, the recognition of real RMS velocity points achieves high picking rates on all three synthetic datasets, which are all above 98\%. On the medium SNR field dataset A, our method can also guarantee a VMAE of 33.980m/s and a VMRE of no more than 1\%. Second, we visualize the picking results of datasets S5 and A. As shown in Figure~\ref{fig:PWRwMPick-1} and Figure~\ref{fig:PWRwMPick-2}, our picking results are very close to the real velocity curve of S5 and the manually picked velocity curve of the field dataset A. We also conduct the NMO correction based on our picking velocity curve and real velocity curve or manually picking velocity curve, as shown in Figure~\ref{fig:NMOGth-1}, \ref{fig:NMOGthGT-1} and  Figure~\ref{fig:NMOGth-2}, \ref{fig:NMOGthGT-2}. The NMO correction results of UEL and labels are almost the same, and both of them can flatten the gather. In Figure~\ref{fig:stk-2560-A}-\ref{fig:stk-2840-M}, we apply the velocities picked by UEL and experts to generate the automatic picking by UEL and labels to generate the stack section of NMO CMP gather. On the overall formation structure of the stack section, the automatic results of UEL are comparable to manual results.

\begin{table}[bt]
\caption{The performance of UEL on four datasets: S1-S3 are the synthetic datasets, and A is a field dataset.}
\centering
\resizebox{0.5\linewidth}{!}{
\begin{tabular}{ccccc}
    \toprule
    Test Set & VMAE   & VMRE (\%)  & PR(\%)      & MD     \\ \hline
    S1      & 9.875  & 0.302 & 100.000 & 9.870  \\
    S3      & 18.227 & 0.593 & 100.000 & 18.331 \\
    S5      & 45.274 & 1.441 & 98.436  & 37.332 \\
    A       & 33.980 & 0.987 & -  & - \\ 
    \bottomrule
    \end{tabular}}
\label{tab:UELTest}
\end{table}

\subsection{Comparative Test}
To verify the advantages of UEL on field surveys, we reproduce two popular unsupervised methods based on the clustering method\citep{bin2019machine} and a supervised method based on deep CNN \citep{ma2018automatic} to compare with our UEL. Concretely, for clustering methods, we set the hyperparameters $K$ in K-Means as 15, $eps$, and the minimum sample number in DBSCAN as 50 and 3, respectively, and the other settings remain the same as in \citep{bin2019machine}. For the CNN method, the training settings all refer to the settings in \citep{ma2018automatic}. 
To test the feasibility of the algorithm in practical applications, we test these four methods on field dataset A with medium SNR. The experimental results are summarized in Table~\ref{tab:comp}. It can be seen that UEL far surpasses popular clustering-based picking methods, and improves the picking accuracy to a very objective level. Even compared to the supervised CNN model, our method achieves better performance. In addition, CNN models generally take at least 1-2 hours to train the model, while our method does not require this training phase, which greatly shortens the seismic data processing cycle.

\begin{table}[bt]
\caption{The test results of the comparative experiment on the field dataset A}
\centering
\resizebox{0.5\linewidth}{!}{
    \begin{tabular}{lcc}
    \toprule
    Method        & VMAE            & VMRE (\%)      \\
    \hline
    DBSCAN\cite[]{bin2019machine}        & 139.760         & 4.278          \\
    K-Means\cite[]{bin2019machine}     & 142.495         & 3.884          \\
    CNN\cite[]{ma2018automatic}           & 45.037          & 1.395          \\ \hline
    \textbf{Ours} & \textbf{33.980} & \textbf{0.987} \\
    \bottomrule
    \end{tabular}}
\label{tab:comp}
\end{table}

\subsection{Robustness to Noise}
\begin{figure}[bt]
\centering
\includegraphics[width=0.6\textwidth]{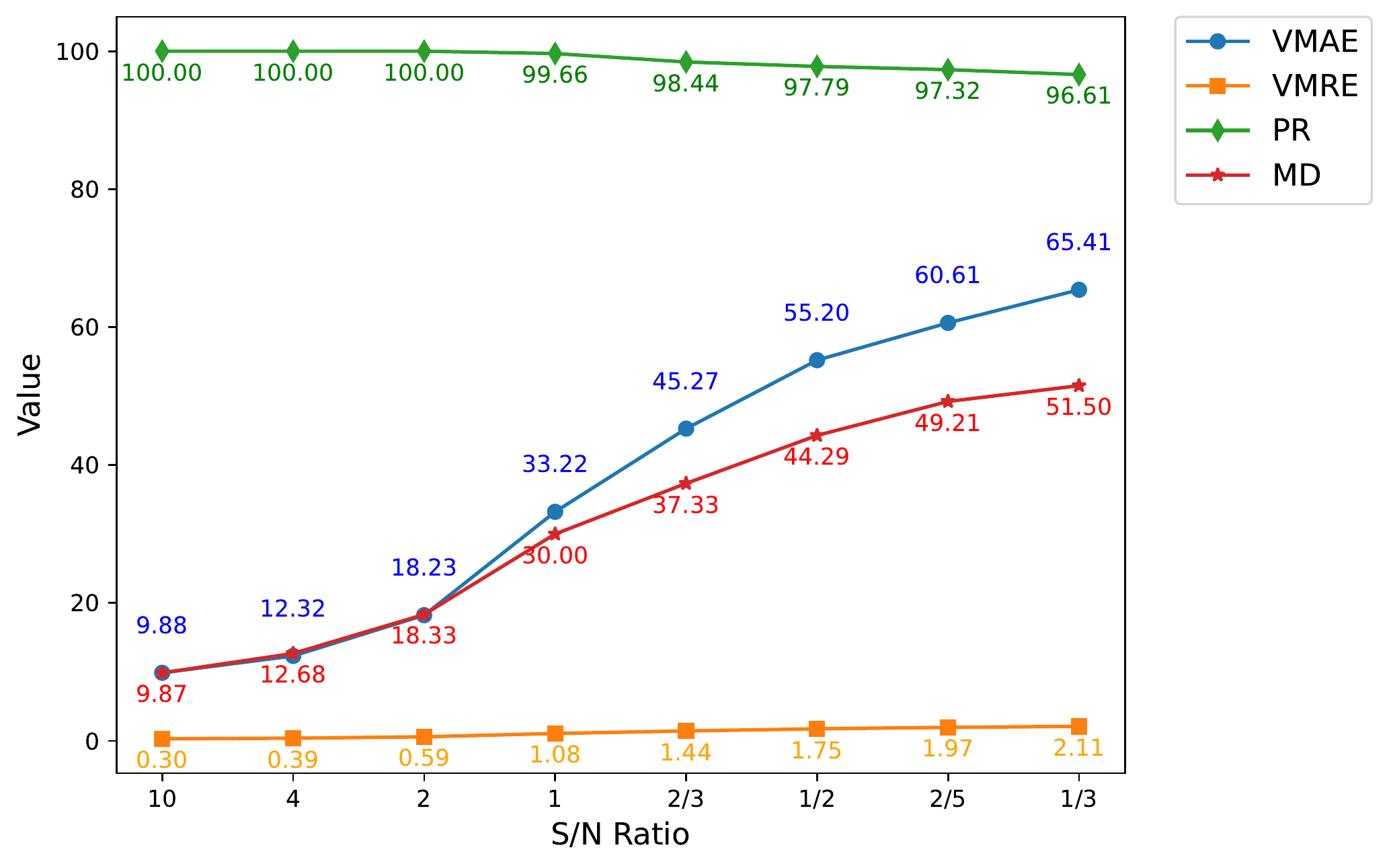}
\caption{Quantization test results of UEL on eight synthetic datasets under different SNRs.}
\label{fig:SNRTest}
\end{figure}

\begin{figure}
    \centering
    \subfloat[]{\includegraphics[height=1.7in]{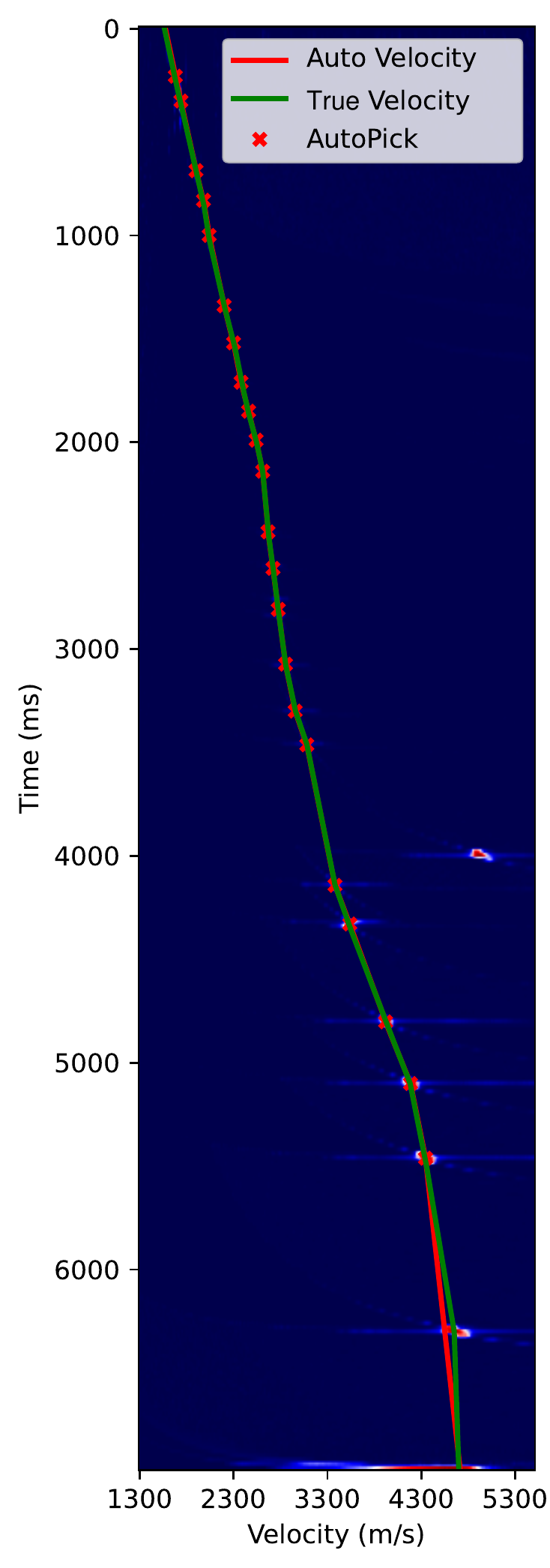}\label{fig:snr-Ep-1}}
    \hfil
    \subfloat[]{\includegraphics[height=1.7in]{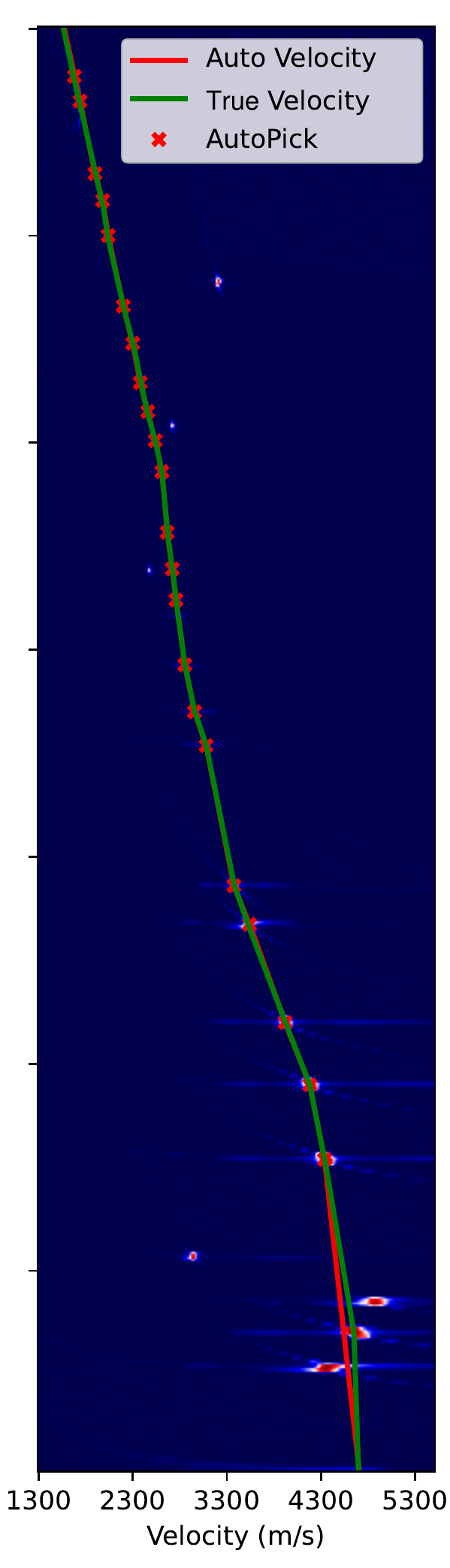}\label{fig:snr-Ep-2}}
    \hfil
    \subfloat[]{\includegraphics[height=1.7in]{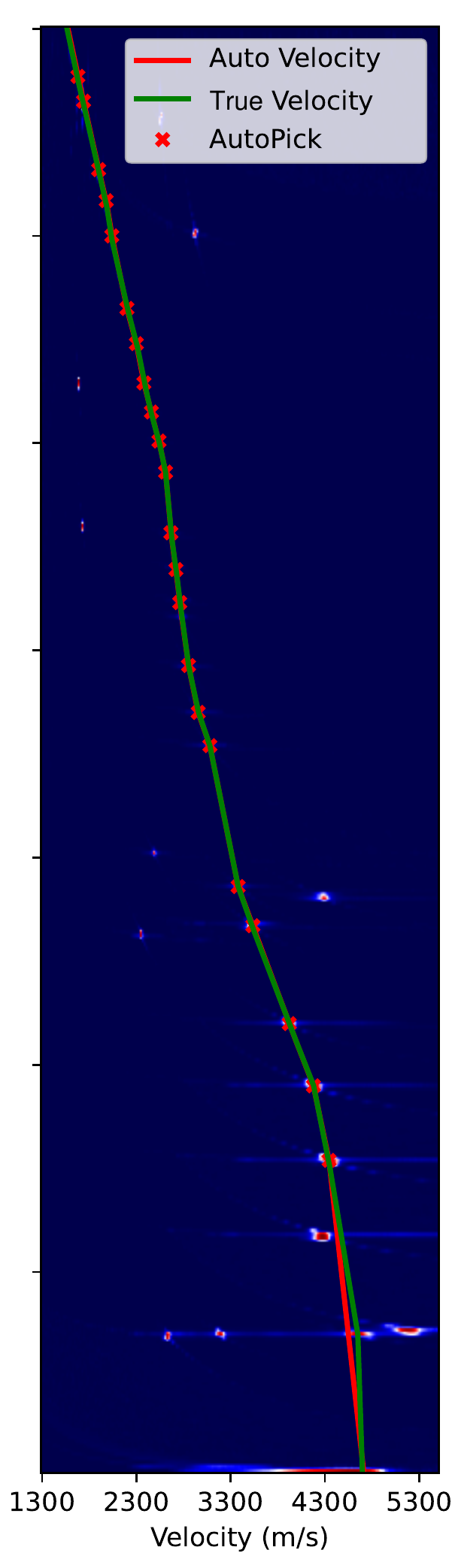}\label{fig:snr-Ep-3}}
    \hfil
    \subfloat[]{\includegraphics[height=1.7in]{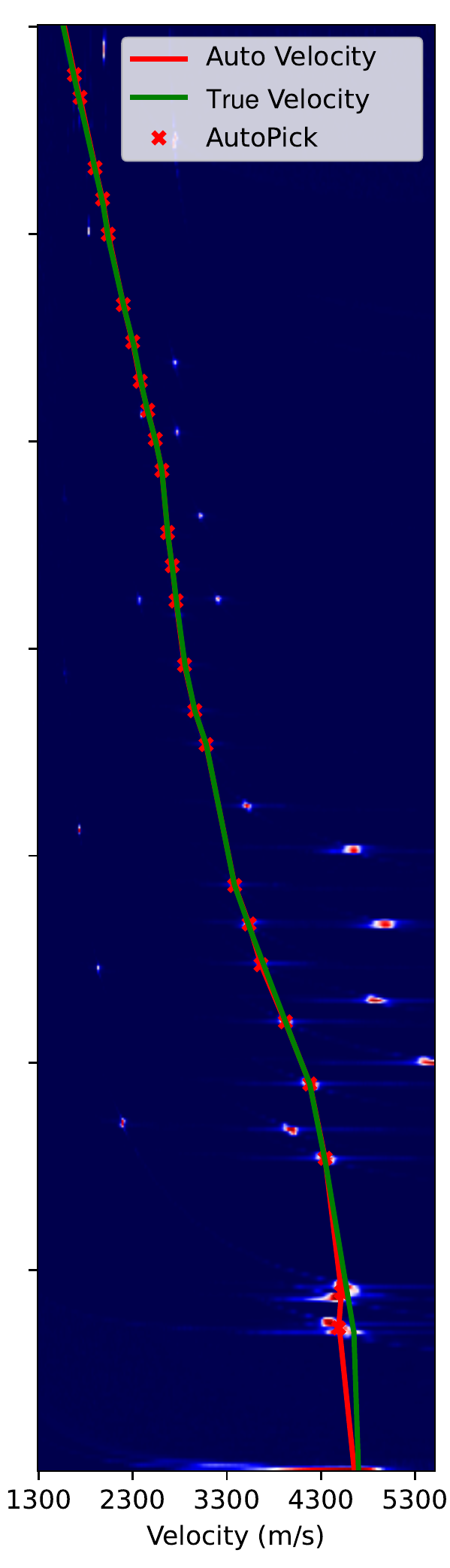}\label{fig:snr-Ep-4}}
    \hfil
    \subfloat[]{\includegraphics[height=1.7in]{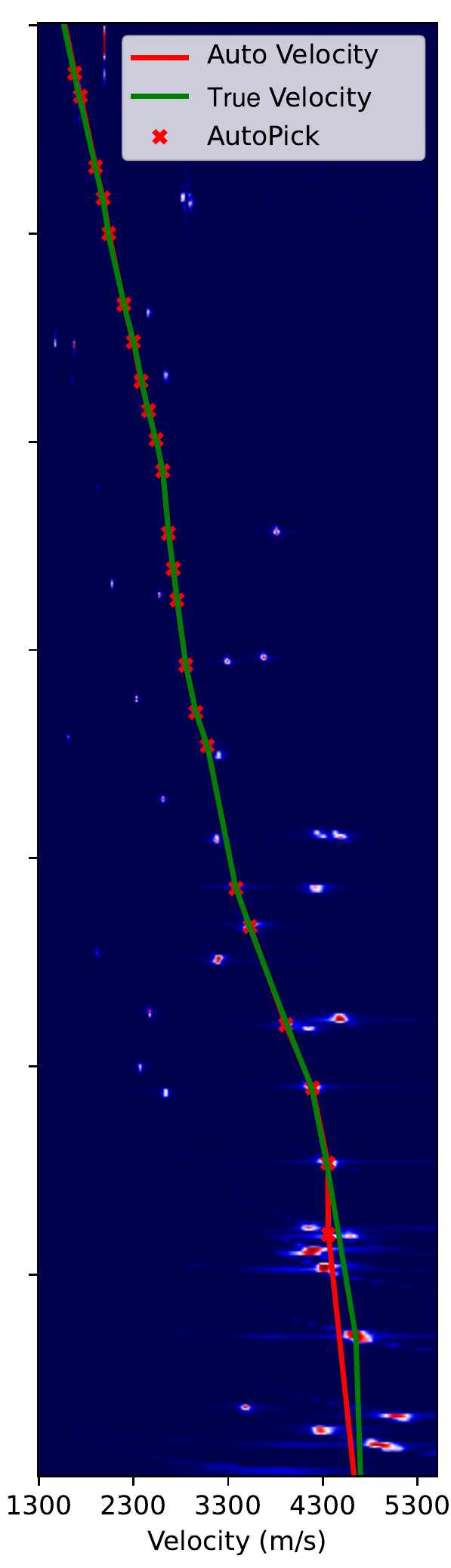}\label{fig:snr-Ep-5}}
    \hfil
    \subfloat[]{\includegraphics[height=1.7in]{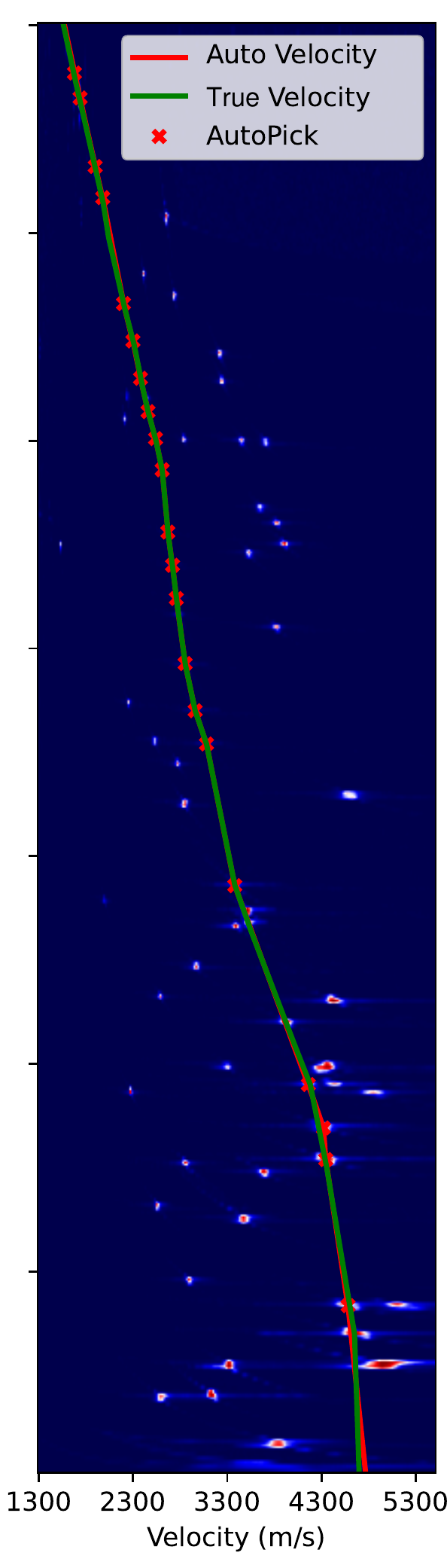}\label{fig:snr-Ep-6}}
    \hfil
    \subfloat[]{\includegraphics[height=1.7in]{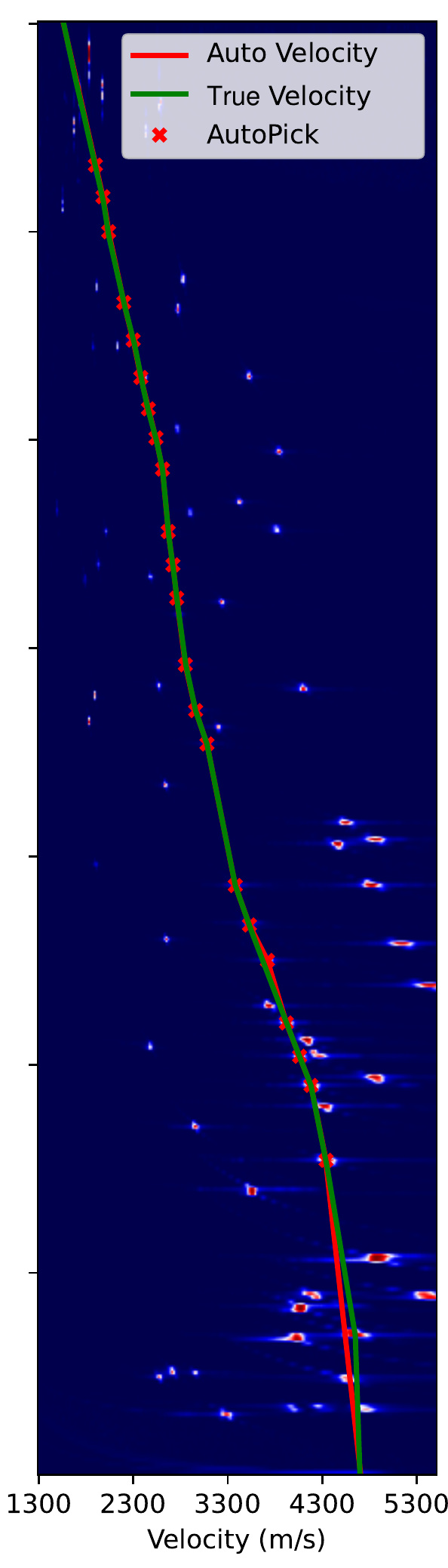}\label{fig:snr-Ep-7}}
    \hfil
    \subfloat[]{\includegraphics[height=1.7in]{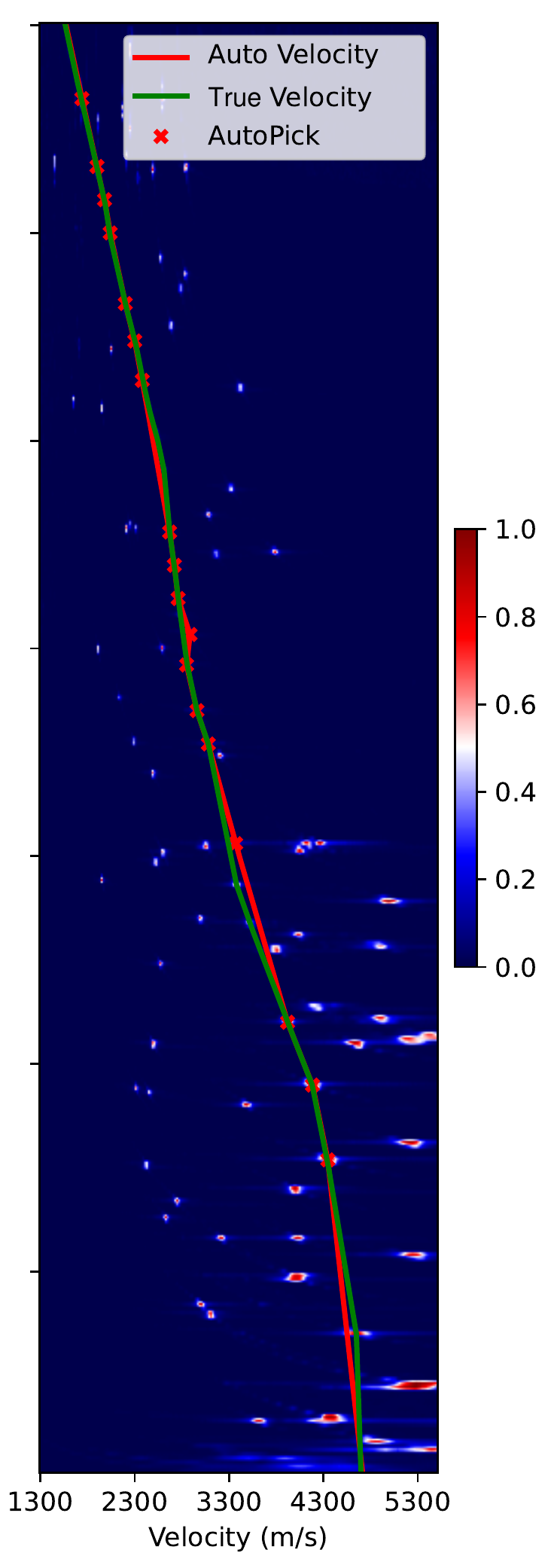}\label{fig:snr-Ep-8}}
    \caption{The picking results of UEL under different SNRs. (a)-(h) are the gained spectrum with velocity curves of S1-S8, respectively. The red crosses are the picking points of UEL. The red curve and the green curve are the velocity curves of UEL and label, respectively.}
    \label{fig:SNRTest2}
\end{figure}

To further explore the noise resistance of our method, we test UEL on eight synthetic datasets with different SNRs. As Figure~\ref{fig:SNRTest} demonstrates, metrics of VMAE, VMRE, and MD increase with decreasing SNR, but even at the lowest SNR of 1/3, we can still achieve VMAE of 65.41m/s, VMRE of 2.11\% and MD of 51.50m/s. PR decreases with decreasing SNR, and all of PRs are above 95\%, which means that most of the real RMS velocity points can be recognized. Figure \ref{fig:SNRTest2} visualizes the picking results of UEL on S1-S8. When the SNR is large, the velocity picking of all temporal depths can be picked up well, as shown in Figure~\ref{fig:snr-Ep-1}-\ref{fig:snr-Ep-3}. As the SNR decreases, the velocity pickup in the middle and deep layers starts to become unstable, but the velocity of the shallow layers can still be picked up well, as shown in Figure~\ref{fig:snr-Ep-4}-\ref{fig:snr-Ep-8}.

\subsection{Ablation Study}
To study the main techniques of UEL, we remove the spectrum gain method, near reference velocity, seed reference velocity, and interval velocity constraint from the UEL, respectively, as shown in Table~\ref{tab:ablation}. Specifically, we first take out two gain modules for the near spectra and current spectrum (named Gain Method in Figure~\ref{fig:MainFlow}), and denote this sub-model as w/o Spectrum Gain, as shown in Table~\ref{tab:ablation}.
After that, the three main parts of the ensemble method are studied. We first do away with the reference velocity of near spectra, represented by w/o Near Velocity, which means that $\text{D}(x, V_{RN}) < w$ is removed in Eq. \ref{eq:RefCons}. Second, removing the reference velocity of seed spectra, denoted as w/o Seed Velocity, which means that $\text{D}(x, V_{RS}) < w$ is removed in Eq. \ref{eq:RefCons}. Finally, we study the necessity of interval velocity constraint, represented by w/o Velocity Constraint, which means that the ensemble method only uses the CA in Eq. \ref{eq:RefCons} to select the reasonable RMS velocity points as the final output.
Concretely, we test these sub-models and the full model UEL on datasets A and S5, as shown in Table~\ref{tab:ablation}, which indicates that the full model UEL outperforms these four sub-models on both datasets. Therefore, we conclude that four important modules can significantly improve the accuracy of velocity picking.

\begin{table}[bt]
\caption{Ablation Study: The effect of four core modules of UEL.}
\centering
\resizebox{0.7\linewidth}{!}{
    \begin{tabular}{lcccc}
    \toprule
    \multirow{2}{*}{Model   Variants} & \multicolumn{2}{c}{\underline{\qquad\qquad Dataset A \qquad\qquad}}           & \multicolumn{2}{c}{\underline{\qquad\qquad Dataset S5 \qquad\qquad}} \\ 
                                      & VMAE            & VMRE (\%)           & VMAE            & VMRE (\%)           \\ \hline
    w/o Spectrum Gain                 & 37.237          & 1.044          & 73.861          & 2.361          \\
    w/o Near Velocity                 & 35.860          & 1.082          & 81.415          & 2.606          \\
    w/o Seed Velocity                 & 53.055          & 1.513          & 85.021          & 2.697          \\
    w/o Velocity Constraint           & 37.952          & 1.132          & 67.510          & 2.128          \\ \hline
    \textbf{Ours}                     & \textbf{33.980} & \textbf{0.987} & \textbf{45.274} & \textbf{1.441} \\ \bottomrule
    \end{tabular}}
\label{tab:ablation}
\end{table}

\section{Conclusions}
In this paper, we present an unsupervised ensemble learning method named UEL for automatically picking RMS velocity from the velocity spectrum. Main conclusions can be drawn from this work, as follows, (1) the information fusion technique of UEL takes full advantage of the both information of the neighbor velocity spectra and seed points, and it greatly improves the picking accuracy of the velocity spectra with medium and low SNRs; (2) compared with the current popular cluster-based unsupervised picking methods and deep CNN-based supervised picking method, UEL can achieve higher picking accuracy on the field dataset with medium SNR; (3) UEL has the strong anti-noise ability, and produces reliable pickup results even at low SNR. 

\section*{acknowledgements}
The research is supported by the National Key Research and Development Program of China under grant 2020AAA0105601, the National Natural Science Foundation of China under grant 61976174 and 61877049.

\section*{conflict of interest}
The  authors  declare  that  they  have  no  known  competing  financial  interests  or  personal relationships that could have appeared to influence the work reported in this paper.

\section*{data availability statement}
The field data set in this study cannot be released under the confidentiality agreement.
\bibliography{sample}



\end{document}